\documentclass[twoside]{article}
\pdfoutput=1

\PassOptionsToPackage{numbers, compress}{natbib}

\usepackage[accepted]{aistats2023}
\usepackage[round, authoryear]{natbib}

\usepackage{hyperref}       
\usepackage{url}            
\usepackage{nicefrac}
\usepackage{xcolor}
\usepackage{graphicx}
\usepackage{subcaption}
\usepackage{hyperref}
\usepackage{booktabs}       
\usepackage{pgfplots}
\usepackage{wrapfig}
\usepackage{amsmath}
\usepackage{amsthm}
\usepackage{cancel}
\usepackage{multirow}
\usepackage{makecell}
\usepackage{algorithm}
\usepackage{algpseudocode}

\usepackage{tikz}
\usetikzlibrary{positioning,fit,calc}
\usepgfplotslibrary{groupplots}
\pgfplotsset{width=7.5cm,compat=1.15, tick label style={font=\small}}

\newcommand\affiliations[1]{%
  \begin{NoHyper}%
\renewcommand\thefootnote{}\footnote{#1}%
  \addtocounter{footnote}{-1}%
  \end{NoHyper}%
}

\usepackage{xspace}

\theoremstyle{plain}

\theoremstyle{definition}

\theoremstyle{remark}

\usepackage{amssymb}
\usepackage{pifont}
\makeatletter
\DeclareRobustCommand\onedot{\futurelet\@let@token\@onedot}
\def\@onedot{\ifx\@let@token.\else.\null\fi\xspace}

\def\eg{\emph{e.g}\onedot} 
\def\ie{\emph{i.e}\onedot}

\usepackage{amsmath,amsfonts,bm}

\newcommand{\Real}{\mathbb{R}}

\def\eqref#1{equation~\ref{#1}}

\def\1{\bm{1}}

\def\rvb{{\mathbf{b}}}

\def\rvs{{\mathbf{s}}}

\def\rvx{{\mathbf{x}}}
\def\rvy{{\mathbf{y}}}
\def\rvz{{\mathbf{z}}}

\DeclareMathAlphabet{\mathsfit}{\encodingdefault}{\sfdefault}{m}{sl}
\SetMathAlphabet{\mathsfit}{bold}{\encodingdefault}{\sfdefault}{bx}{n}

\def\sP{{\mathbb{P}}}
\def\sQ{{\mathbb{Q}}}
\def\sR{{\mathbb{R}}}

\newcommand{\KL}{{\mathrm{KL}}}
\renewcommand{\H}{{\mathrm{H}}}
\newcommand{\I}{{\mathrm{I}}}

\makeatother

\makeatother

\begin{document}

%

%

\twocolumn[

\aistatstitle{Spread Flows for Manifold Modelling}

\aistatsauthor{Mingtian Zhang \And Yitong Sun \And  Chen Zhang \And Steven McDonagh }

\aistatsaddress{ University College London$^*$ \And Huawei Noah’s Ark Lab \And Huawei Noah’s Ark Lab \And Huawei Noah’s Ark Lab} ]

\affiliations{
$^*$This work was done during an internship in Huawei Noah's Ark Lab. Correspondence to: Mingtian Zhang 
\href{mailto:m.zhang@cs.ucl.ac.uk}{m.zhang@cs.ucl.ac.uk}.
}

\begin{abstract}
Flow-based models typically define a latent space with dimensionality identical to the observational space. In many problems, however, the data does not populate the full ambient data space that they natively reside in, rather inhabiting a lower-dimensional manifold. In such scenarios, flow-based models are unable to represent data structures exactly as their densities will always have support off the data manifold, potentially resulting in degradation of model performance. 
To address this issue, we propose to learn a manifold prior for flow models that leverage the recently proposed \emph{spread divergence} towards fixing the crucial problem; the KL divergence 
and maximum likelihood estimation are ill-defined for manifold learning. In addition to improving both sample quality and representation quality,
an auxiliary benefit enabled by our approach is the ability to  identify the intrinsic dimension of the manifold distribution.
\end{abstract}

\section{Introduction}

Normalizing flows~\cite{rezende2015variational} have shown considerable potential for the task of modelling and inferring expressive distributions through the learning of well-specified probabilistic models. Recent progress in this area has defined a set of general and extensible structures, capable of representing highly complex and multimodal distributions, see~\cite{kobyzev2020normalizing} for a detailed overview. 
Specifically, assume an absolutely continuous~\footnote{The distribution is a.c.~with respect to the Lebesgue measure, so it has a density function, see \cite{durrett2019probability}.}~(a.c.)~random variable (r.v.)~$Z$ with distribution $\sP_Z$  and probability density $p_Z(z)$. We can transform $Z$ to get a r.v.~$X$: $X=f(Z)$, where $f:\Real ^D\rightarrow\Real ^D$ is an invertible function with inverse $f^{-1}=g$, so  $X$ has a (log) density function $p_X(x)$ with  the following form
\begin{equation}
 \log p_X(x)=\log p_Z(g(x))+\log\left|\det\left( \frac{\partial g}{\partial x}\right)\right|,
\end{equation}
where $\log\left|\det\left( \frac{\partial g}{\partial x}\right)\right|$ is the log determinant of the Jacobian matrix. We call $f$ (or $g$) a \emph{volume-preserving} function if the log determinant is equal to 0. Training of flow models typically makes use of MLE. We assume the data random variable $X_d$ with distribution $\sP_d$ is a.c.~and has density $p_d(x)$. In addition to the well-known connection between MLE and minimization of the KL divergence $\KL(p_d(x)||p_X(x))$ in $X$ space (see Appendix~\ref{app:mle:kl} for details), MLE is also equivalent to minimizing the KL divergence in $Z$ space, due to the KL divergence invariance under invertible transformations~\cite{yeung2008information,papamakarios2019normalizing}. 
Specifically, we define $Z_\sQ: Z_\sQ=g(X_d)$ with distribution $\sQ_Z$ and density function\footnote{Since we assume $X_d$ is a.c., $Z_\sQ: Z_\sQ=g(X_d)$ is also a.c.~with a bijiective mapping $g(\cdot)$ and thus $Z_\sQ$ allows a density function.} $q(z)$, the KL divergence in $Z$ space  $\KL(q(z)||p(z))$ can be written as
\begin{align}\label{eq:klz}
    -\int p_d(x)\left(\log p_Z\left(g(x)\right)+\log \left|\det\left(\frac{\partial g}{\partial x}\right)\right|\right)dx,
\end{align}
with constant term discarded, the full derivation can be found in Appendix \ref{app:mle:kl}. Since we can only access samples $x_1,x_2,\ldots,x_N$ from $p_d(x)$, we approximate the integral by Monte Carlo sampling
 \begin{equation}\label{eq:klz:mle}
     \KL(q(z)||p(z))\approx-\frac{1}{N}\sum_{n=1}^N\log p_X(x_n)+const..
\end{equation}
We highlight the connection between MLE and KL divergence minimization in $Z$ space for flow models. The prior distribution $p(z)$ is usually chosen to be a $D$-dimensional Gaussian distribution.

Further to this, we note that if data lies on a lower-dimensional manifold, and thus does not populate the full ambient space, then the estimated flow model will necessarily have mass lying off the data manifold, which may result in under-fitting and poor generation qualities. Contemporary flow-based approaches may make for an inappropriate representation choice in such cases. Formally, when data distribution $\sP_d$ is singular, \eg a measure on a low dimensional manifold, then $\sP_d$ or the induced latent distribution $\sQ_Z$ will no longer emit valid density functions. In this case, the KL divergence and MLE  in~\eqref{eq:klz} are typically not well-defined under the considered flow model assumptions. This issue brings theoretical and practical challenges that we discuss in the next section.

\section{Flow Models for Manifold Data
\label{sec:manifold:data}}

\begin{figure*}[t]
\centering
\begin{subfigure}{0.3\textwidth}
\centering
\includegraphics[width=0.8\textwidth]{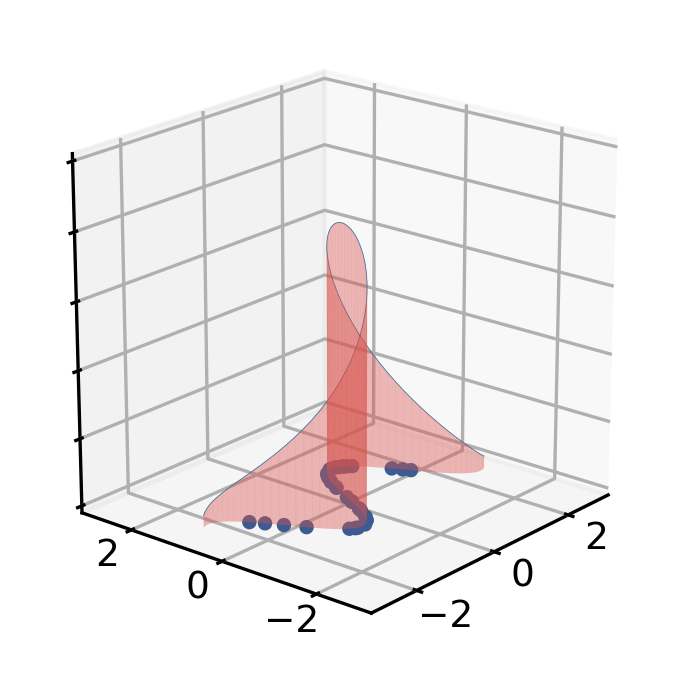} 
\caption{$\sP_d$}
\label{fig:sindatadis}
\end{subfigure}
\begin{subfigure}{0.3\textwidth}
\centering
\includegraphics[width=0.8\textwidth]{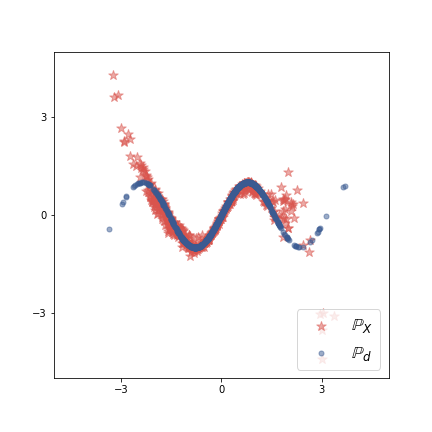} 
\caption{$X$ space}
\label{fig:sin:samples}
\end{subfigure}
\begin{subfigure}{0.3\textwidth}
\centering
\includegraphics[width=0.8\textwidth]{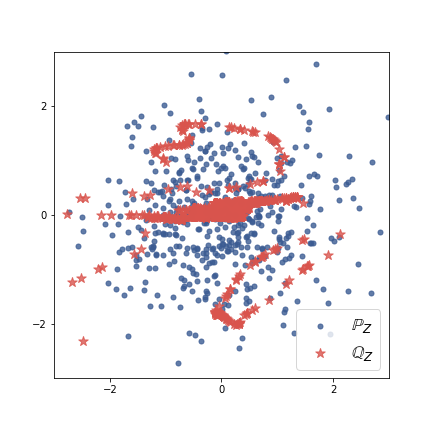}
\caption{$Z$ space}
\label{fig:sin:latent}
\end{subfigure}
\caption{Samples and latent visualization from a flow based model with a fixed Gaussian prior where the intrinsic dimension is strictly lower than the true dimensionality of the data space.}
\vspace{-0.2cm}
\end{figure*}

We assume a data sample $\rvx \sim \sP_d$ to be a $D$ dimensional vector $\rvx \in \sR^D$ and define the ambient dimensionality of $\sP_d$, denoted by $\texttt{Amdim}(\sP_d)$, to be $D$. However for many datasets of interest, \eg natural images, the data distribution $\sP_d$ is commonly believed to be supported on a lower dimensional manifold~\cite{beymer1996image}. We assume the dimensionality of the manifold to be $K$ where $K < D$, and define the intrinsic dimension of $\sP_d$, denoted by $\texttt{Indim}(\sP_d)$, to be the dimension of this manifold. Figure~\ref{fig:sindatadis} provides an example of this setting where $\sP_d$ is a 1D distribution in 2D space. Specifically, each data sample $\rvx\sim \sP_d$ is a 2D vector $\rvx=\{x_1,x_2\}$ where $x_1\sim \mathcal{N}(0,1)$ and $x_2=\sin(2x_1)$. Therefore, this example results in $\texttt{Amdim}(\sP_d)=2$ and $\texttt{Indim}(\sP_d)=1$.

In flow-based models, function $f$ is constructed such that it is both bijective and differentiable. When the prior $\sP_Z$ is a distribution whose support is $\sR^D$ (\eg Multivariate Gaussian distribution), the marginal distribution $\sP_X$ will also have support $\sR^D$ and $\texttt{Amdim}(\sP_X)=\texttt{Indim}(\sP_X)=D$. When the support of the data distribution lies on a $K$-dimensional manifold and $K < D$, $\sP_d$ and $\sP_X$ are constrained to have different support. That is, the intrinsic dimensions of $\sP_X$ and $\sP_d$ are always different; $\texttt{Indim}(\sP_X)\neq \texttt{Indim}(\sP_d)$. In this case it is impossible to learn a model distribution $\sP_X$ identical to the data distribution $\sP_d$. Nevertheless, flow-based models have shown strong empirical success in real-world problem domains such as the ability to generate high-quality and realistic images~\cite{kingma2018glow}. Towards investigating the cause, and explaining this disparity between theory and practice, we employ a toy example to provide intuition for the effects and consequences resulting from model and data distributions that possess differing intrinsic dimensions.

Consider the toy dataset introduced previously; a 1D distribution lying in a 2D space (Figure \ref{fig:sindatadis}). The prior density $p(z)$ is a standard 2D Gaussian $p(z)=\mathcal{N}(0,I_Z)$ and the function $f$ is a non-volume preserving flow with two coupling layers (see Appendix \ref{app:network}). In Figure~\ref{fig:sin:samples} we plot samples from the flow model; the sample $\rvx$ is generated by first sampling a 2D datapoint $\rvz \sim \mathcal{N}(0,I_Z)$ and then letting $\rvx=f(\rvz)$.  Figure~\ref{fig:sin:latent} shows samples from the prior distributions $\sP_Z$ and $\sQ_Z$. $\sQ_Z$ is defined as the transformation of $\sP_d$ using the bijective function $g$, such that $\sQ_Z$ is constrained to support a 1D manifold in 2D space, and $\texttt{Indim}(\sQ_Z)=\texttt{Indim}(\sP_d)=1$. Training of $\sQ_Z$ to match $\sP_Z$ (which has intrinsic dimension 2), can be seen to result in ``curling up'' of the manifold in the latent space, contorting it towards satisfying a distribution that has intrinsic dimension 2 (see Figure~\ref{fig:sin:latent}). This ill-behaved phenomenon causes several potential problems for contemporary flow models:
\begin{enumerate}

    \item \emph{Poor sample quality.} Figure~\ref{fig:sin:samples} shows examples where incorrect assumptions result in the model generating poor samples. 
    
    \item \emph{Low quality data representations.} The discussed characteristic that results in ``curling up'' of the latent space may cause representation quality degradation. 
    
    \item \emph{Inefficient use of network capacity}. Neural network capacity is spent on contorting the distribution $\sQ_Z$ to satisfy imposed dimensionality constraints. 
    
\end{enumerate}

A natural solution to the problem of intrinsic dimension mismatch is to select a prior distribution $\sP_Z$ with the same dimensionality as the intrinsic dimension of the data distribution such that: $\texttt{Indim}(\sP_Z)=\texttt{Indim}(\sP_d)$. However, since the $\texttt{Indim}(\sP_d)$ is unknown, one option is to instead learn it from the data. In the next section, we  introduce an approach that enables us to learn $\texttt{Indim}(\sP_d)$.

\section{Learning a Manifold Prior\label{sec:identification}}
\begin{figure}
    \centering
\includegraphics[scale=0.5]{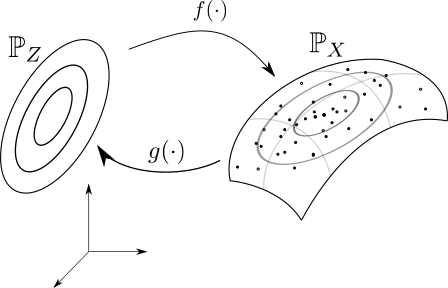}
    \caption{This figure gives an overview of our method. For data (black dots) that lie on a 2D manifold in 3D space, we want to use a model $\sP_X$ with $\texttt{Indim}(\sP_X)=2$. Therefore, we learn a prior $\sP_Z$ that is a 2D Gaussian in 3D space and an invertible function $f$ which maps from $\sP_Z$ to $\sP_X$.}\label{fig:overview}
\end{figure}
Consider a data vector $\rvx\in\sR^{D}$, then a flow-based model prior $\sP_Z$ is usually given by a $D$-dimensional Gaussian distribution or alternative simple distribution that is also absolutely continuous (a.c.)~in $\sR^D$.
Therefore, the intrinsic dimension $\texttt{Indim}(\sP_Z)=D$. To allow a prior to have an intrinsic dimension strictly less than $D$, we let $\sP_Z$ be a `\emph{generalized Gaussian}'\footnote{We use the generalized Gaussian to include the case that the covariance $AA^T$ is not full rank.} distribution $\mathcal{GN}(0,AA^T)$,
where $\rvz\in\sR^D$ and $A$ is a $D{\times}D$ lower triangular matrix with $D(D+1)/2$ parameters, such that $AA^T$ is constrained to be a positive semi-definite matrix. When $AA^T$ has full rank $D$, then $\sP_Z$ is a (non-degenerate) multivariate Gaussian  on $\sR^D$. When $\texttt{Rank}(AA^T) = K$ and 
$K < D$, then $\sP_Z$ will degenerate to a Gaussian supported on a $K$-dimensional manifold, such that the intrinsic dimension $\texttt{Indim}(\sP_Z)=K$\footnote{By the definition of the manifold~\cite{lee2013smooth}, the intrinsic dimension of a 
manifold is equal or smaller than its ambient dimension.}.
Figure~\ref{fig:overview} illustrates a sketch of this scenario. In practice, we initialize $A$ to be an identity matrix, thus $AA^T$ is also an identity and $\sP_Z$ is initialized as a standard Gaussian.  We want to highlight that the classic flow method with a standard Gaussian prior is a special case with $A=I$ in our model and the additional free parameters in matrix $A$ allow the model to capture the manifold property of the target data distribution.

\textbf{Identification of the Intrinsic Dimension} A byproduct of our model is that the intrinsic dimension of the data manifold can be identified. 
When the model matches the true data distribution $\sP_X=\sP_d$, the supports of $\sP_X$ and $\sP_d$  will also have the same intrinsic dimension: $\texttt{Indim}(\sP_X)=\texttt{Indim}(\sP_d)$. The flow function $f$, and its inverse $g=f^{-1}$, are bijective and continuous so $f$ is a diffeomorphism~\cite{kobyzev2020normalizing}. Due to the \emph{invariance of dimension} property of diffeomorphisms~\cite{lee2013smooth}, the manifold that supports $\sP_Z$ will have the same dimension as the manifold that supports $\sP_X$ thus
    \begin{equation}
    \texttt{Indim}(\sP_Z)=\texttt{Indim}(\sP_X)=\texttt{Indim}(\sP_d).
\end{equation}
Since the intrinsic dimension of $\sP_Z$ is equal to the rank of the matrix $AA^T$, we can calculate $\texttt{Rank}(AA^T)$ by counting the number of non-zero eigenvalues of the matrix $AA^T$. This allows for identification of the intrinsic dimension of the data distribution as 
\begin{equation}
    \texttt{Indim}(\sP_d)=\texttt{Rank}(AA^T).
\end{equation}

\textbf{Dimension Reduction} We would also like to conduct dimension reduction using the learned manifold flow model. For $\sQ_Z$ with $\texttt{Amdim}(\sQ_Z){=}D$ and $\texttt{Indim}(\sQ_Z){=}K$,  we first conduct an eigen-value decomposition of the $D\times D$ matrix $AA^T$: $AA^T=\mathbf{E}\Lambda \mathbf{E}^T$, where $\mathbf{E}=[e_1,\cdots,e_D]$ contains all the eigen-vectors and $\Lambda=\mathrm{diag}(\lambda_1,\cdots,\lambda_D)$.
When $\texttt{Rank}(AA^T)=K \leq D$, there exist $K$ eigenvectors with positive eigenvalues. We select the first $K$ eigenvectors and form the matrix $\mathbf{E}=[\mathbf{e}^{1},\ldots,\mathbf{e}^K]$ with dimension $D{\times}K$. We then transform each data sample $\rvx\in\mathbb{R}^D$ into $Z$ space: $\rvz=g(\rvx)$, such that $\rvz\in\mathbb{R}^D$. Finally, a linear projection is carried out $\rvz^{proj}=\rvz\mathbf{E}$ to obtain the lower dimensional representation $\rvz^{proj}\in \mathbb{R}^K$. This procedure can be seen as a \emph{nonlinear PCA}, where the nonlinearity is learned by the inverse flow function $g$.

\textbf{Sample from the Manifold Flow} To sample from the flow model with $\texttt{Indim}(\sP_X)$, we can first take a sample
$\epsilon$ from a standard $K$-dimensional Gaussian distribution and let $\mathrm{z}'=\mathbf{E}_K\sqrt{\Lambda_K}$ where $\mathbf{E}_K$ are the first $K$ eigenvectors and  
$\Lambda_K$ is the corresponding eigen-values. When $K=D$, the linear matrix $\mathbf{E}_K\sqrt{\Lambda_K}$ will be equivalent to the Cholesky decomposition solution of the matrix $AA^T$. We can then let $\mathrm{x}'=f(\mathrm{z}')$ as the sample from the model.

\section{Addressing Ill-defined KL divergence \label{sec:spread}}
When $\texttt{Rank}(AA^T) < D$, the degenerate covariance $AA^T$ is no longer invertible and we are unable to evaluate the density value of $p(\rvz)$ for a given random vector $\rvz$. Furthermore, when the data distribution $\sP_d$ is supported on a 
 $K$-dimensional manifold, $\sQ_Z$ will also be supported on a $K$-dimensional manifold and no longer has a valid density function. Using \eqref{eq:klz} to train the flow 
model then becomes impossible as the KL divergence between $\sP_Z$ and $\sQ_Z$ is not well defined\footnote{The KL divergence $\KL(\sQ||\sP)$ is well defined when $\sQ$ and $\sP$ have valid densities with common support~\cite{ali1966general}.}. Recent work by~\cite{spread, zhang2019variational} proposed a new family of divergence to address this problem, which we introduce below.

Let $Z_\sQ$ and $Z_\sP$ be two random variables  with distributions $\sQ_Z$ and $\sP_Z$, respectively. The KL divergence between $\sQ_Z$ and $\sP_Z$ is not well defined if $\sQ_Z$ or $\sP_Z$ does not have valid density functions. Let $K$ be an a.c.~random variable that is independent of $Z_\sQ$ and $Z_\sP$ and has density  $p_K$, We define $Z_{\tilde{\sP}}=Z_\sP+K; Z_{\tilde{\sQ}}=Z_\sQ+K$ with distributions $\tilde{\sP}_Z$ and $\tilde{\sQ}_Z$ respectively. Then  $\tilde{\sP}_Z$ and $\tilde{\sQ}_Z$ are a.c. \citep[Theorem 2.1.16]{durrett2019probability} with density functions
\begin{align}\label{eq:spread_density}
\resizebox{0.905\hsize}{!}{$
    q(\tilde{\rvz})=\int_{\rvz} p_K(\tilde{\rvz}-\rvz)d\sQ_Z,\quad p(\tilde{\rvz})=\int_{\rvz} p_K(\tilde{\rvz}-\rvz)d\sP_Z.$}
\end{align}
The \emph{spread KL divergence} between $\sQ_Z$ and $\sP_Z$ as the KL divergence between $\tilde{\sQ}_Z$ and $\tilde{\sP}_Z$ is:
\begin{align}\label{eq:spread_KL_divergence}
    \widetilde{\KL}(\sQ_Z||\sP_Z)\equiv \KL(\tilde{\sQ}_Z||\tilde{\sP}_Z)\equiv \KL\left( q(\tilde{\rvz})||p(\tilde{\rvz})\right).
\end{align}
In this work we let $K$ be a Gaussian with diagonal covariance $\sigma_Z^2I$ to satisfy the sufficient conditions such that $\widetilde{\KL}(\cdot)$ is a valid divergence (see \cite{spread} for details) and has the properties:
\begin{align}
\resizebox{0.905\hsize}{!}{$
\widetilde{\KL}(\sQ_Z||\sP_Z)\geq 0,\quad\widetilde{\KL}(\sQ_Z||\sP_Z)=0\Leftrightarrow \sQ_Z=\sP_Z.$}
\end{align}
Since $\sQ_Z$ and $\sP_Z$ are transformed from $\sP_d$ and $\sP_X$ using an invertible function $g$, we have 
\begin{align}
\sQ_Z=\sP_Z\Leftrightarrow \sP_d=\sP_X.
\end{align}
Therefore, the spread KL divergence can be used to train flow-based models with a manifold prior in order to fit a dataset that lies on a lower-dimensional manifold. 


\subsection{Estimation of the Spread KL Divergence}
As shown in \eqref{eq:spread_KL_divergence}, minimizing $\widetilde{\KL}(\sQ_Z||\sP_Z)$ is equivalent to minimizing $\KL( q(\tilde{\rvz})||p(\tilde{\rvz}))$, which has two terms
\begin{align}\label{eq:spread:kl}
\resizebox{0.85\hsize}{!}{$
\widetilde{\KL}(\sQ_Z||\sP_Z)=\underbrace{\int q(\tilde{\rvz})\log  q(\tilde{\rvz})d\tilde{\rvz }}_{\textbf{Term 1}}-\underbrace{\int q(\tilde{\rvz})\log p(\tilde{\rvz})d\tilde{\rvz}}_{\textbf{Term 2}}$},
\end{align}
where $q(\tilde{\rvz})$ and $p(\tilde{\rvz})$ are defined in~\eqref{eq:spread_density}. We next discuss the estimation  of this objective.

\textbf{Term 1:} We use $\H(\cdot)$ to denote the differential entropy. Term 1 is the denoted as $-\H(Z_{\tilde{\sQ}})$. For a volume-preserving $g$ and an a.c.~$X_d$, the entropy $\H(Z_\sQ)=\H(X_d)$ is independent of the model parameters and can be ignored during training. However, the entropy $\H(Z_{\tilde{\sQ}})=\H(Z_\sQ+K)$ will still depend on $g$, see Appendix \ref{app:an:example} for an example. We claim that when the variance of $K$ is small, the dependency between $\H(Z_{\tilde{\sQ}})$ and the volume preserving function $g$ is weak, thus we can approximate \eqref{eq:spread:kl} by 
disregarding term 1 and this will not adversely affect the training. 

To build intuitions, we  assume $X_d$ is a.c., so $Z_\sQ=g(X_d)$ is also a.c.. Using standard entropic properties~\cite{kontoyiannis2014sumset}, we have the following relationship
\begin{align}\label{eq:entropy:inequality}
    \mathrm{H}(Z_\sQ)\leq\mathrm{H}(Z_\sQ+K)= \mathrm{H}(Z_\sQ)+\mathrm{I}(Z_\sQ+K,K),
\end{align}
where $\mathrm{I}(\cdot,\cdot)$ denotes the mutual information. Since $Z_\sQ$ is independent of function $g$ and if $\sigma_Z^2\rightarrow 0$, then $\I(Z_\sQ+K,K)\rightarrow 0$ (see Appendix \ref{app:ac:entropy} for a proof), the contribution of the $\mathrm{I}(Z_\sQ+K,K)$ term, with respect to training $g$, becomes negligible in the case of small $\sigma_Z^2$. 

Unfortunately, \eqref{eq:entropy:inequality} is no longer valid when $\sP_d$ lies on a manifold since $Z_\sQ$ will be a singular random variable and the differential entropy $\H(Z_\sQ)$ is not defined. In Appendix \ref{app:upper:bound}, we show that leaving out the entropy $\H(Z_{\tilde{\sQ}})$ corresponds to minimizing an \emph{upper bound} of the spread KL divergence. To further explore 
the
contribution of the negative entropy term, we compare 
leaving out $-\H(Z_{\tilde{\sQ}})$ 
with alternatively approximating $-\H(Z_{\tilde{\sQ}})$ during training. In Appendix \ref{app:empirical:evidence}, we discuss an appropriate approximation technique and 
provide empirical evidence which shows that 
ignoring $-\H(Z_{\tilde{\sQ}})$ will not adversely affect the training of our model. Therefore, we make use of volume preserving $g$ and small variance $\sigma^2_Z=1\times 10^{-4}$ in our experiments.

In contrast to other volume-preserving flows, that utilize a fixed prior $\sP_Z$, our method affords additional flexibility by way of allowing for changes to the `volume' of the prior towards matching the distribution of the target data. In this way, our decision to employ volume-preserving flow functions does not limit the expressive power  of the model, in principle. Popular non-volume preserving flow structures, \eg affine coupling flow, may also easily be normalized to become volume preserving, thus further extending the applicability of our approach (see Appendix \ref{app:network}).

\textbf{Term 2:}
The noisy prior $p(\tilde{\rvz})$ is defined to be a degenerate Gaussian $\mathcal{N}(0,AA^T)$, convolved with Gaussian noise $\mathcal{N}(0,\sigma_Z^2I)$, and has a closed form density 
\begin{align}\label{eq:noisy:prior}
p(\tilde{\rvz})=\mathcal{N}(0,AA^T+\sigma_Z^2I).
\end{align}
Therefore, the log density $\log p(\tilde{\rvz})$ is well defined, we can approximate term 2 using Monte Carlo
\begin{align}
    \int q(\tilde{\rvz})\log p(\tilde{\rvz})d\tilde{\rvz}\approx \frac{1}{N}\sum_{n=1}^{N} \log p(\tilde{\rvz}_n), \label{eq:loss:function}
\end{align}
where $ q(\tilde{\rvz})=\int p(\tilde{\rvz}|\rvz)d\sQ_Z$. To sample from $ q(\tilde{\rvz})$, we first get a data sample $\rvx\sim \sP_d$, use function $g$ to get $\rvz=g(\rvx)$ (so $\rvz$ is a sample of $\sQ_Z$) and finally sample $\tilde{\rvz}\sim p(\tilde{\rvz}|\rvz)$. The training details can be found in Algorithm~\ref{alg:cap}.

\renewcommand{\algorithmicrequire}{\textbf{Given:}}
\renewcommand{\algorithmicensure}{\textbf{Model:}}
\begin{algorithm}
\caption{Training Spread Flows}\label{alg:cap}
\begin{algorithmic}
\Require $\mathcal{X}_{train}=\{\rvx_1,\cdots, \rvx_N\} \sim \sP_d$, $\mathcal{N}(0,\sigma_Z^2I)$
\Ensure An inverse flow function $g$, a  lower triangular matrix $A$ initialized as an identity matrix.
\While{not  converge}
\State Sample a data batch  $\mathcal{X}_{B}=\{\rvx_1,\cdots, \rvx_B\} \in\mathcal{X}_{train}$.
\State Transform data $z_b=g(\rvx_b)$ for $\rvx_b\in \mathcal{X}_{B}$.
\State Sample noise $\epsilon_b\sim \mathcal{N}(0,\sigma_Z^2I)$ for each $\rvx_b$.
\State Add noise $\tilde{\rvz}_b=\rvz_b+\epsilon_b$ for each latent representation.
\State Train $g_\theta$ and $A$ using Equation~\ref{eq:loss:function}.
\EndWhile
\end{algorithmic}
\end{algorithm}

\section{Experiments\label{sec:exp}}

Traditional flows assume the data distributions are a.c. and we 
extend this assumption such that the data distribution lies on one continuous manifold. We demonstrate both the effectiveness and robustness of our model by considering firstly (1) data distributions that satisfy our continuous manifold  assumption: toy 2D and 3D data, the \emph{fading square} dataset; and secondly  (2) distributions where our assumption no longer holds: synthesized and real MNIST~\cite{lecun1998mnist}, 
and CelebA~\cite{celeba}.
We use incompressible affine coupling layers, introduced by~\cite{sorrenson2020disentanglement,dinh2016density}, for toy and MNIST experiments and a volume-preserving variation of the Glow structure~\cite{kingma2018glow}. See Appendix~\ref{app:exp} for further experimental details.
%

\subsection{Synthetic Data} 
\begin{figure*}[t]
    \begin{minipage}[t]{\textwidth}
    \centering
    \begin{subfigure}{0.175\linewidth}
    \centering
    \includegraphics[width=1.13\linewidth]{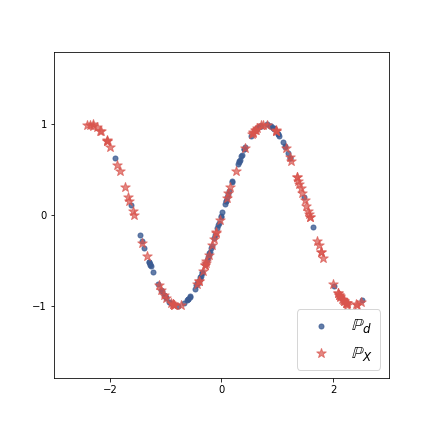}
    \caption{$X$ space }
    \end{subfigure}
    \begin{subfigure}{0.175\linewidth}
    \centering
    \includegraphics[width=1.13\linewidth]{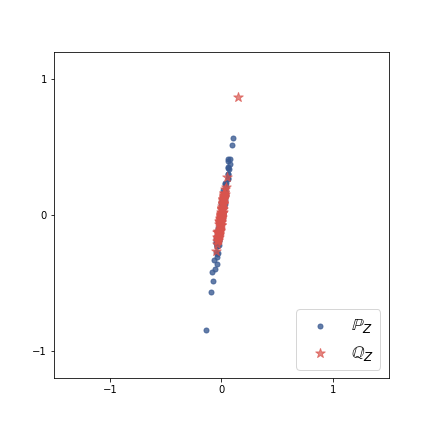}
    \caption{$Z$ space }
    \end{subfigure}
    \begin{subfigure}{0.125\linewidth}
    \centering
    \includegraphics[width=1.03\linewidth]{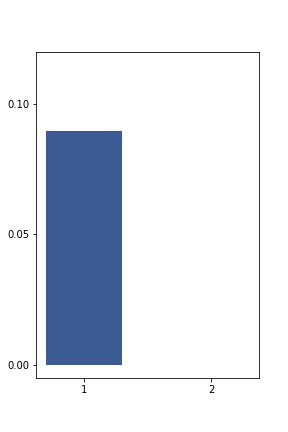}
    \caption{Eigenval.}
    \end{subfigure}
    \begin{subfigure}{0.25\linewidth}
    \centering
    \includegraphics[width=1.13\linewidth]{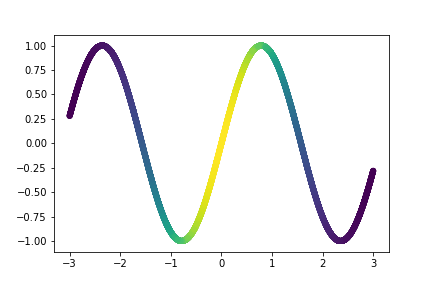}
    \caption{True Density}
    \end{subfigure}
    \begin{subfigure}{0.25\linewidth}
    \includegraphics[width=1.13\linewidth]{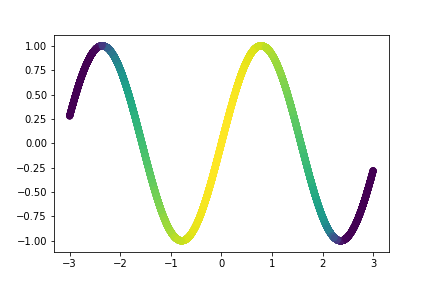}
    \caption{Our Estimation}
    \end{subfigure}

    \caption{(a) shows the samples from $\sP_d$ (blue) and model $\sP_X$ (red). (b) shows the samples from the learned prior $\sP_Z$ (blue) and  $\sQ_Z$ (red). (c) shows the eigenvalues of  $AA^T$. (d) and (e) show the true and the learned density on the manifold. \label{fig:sin}}
    \end{minipage}
    \end{figure*}

\textbf{2D toy data}
We firstly verify our method using the toy data depicted in Figure~\ref{fig:sindatadis}. Figure~\ref{fig:sin} shows model samples, the learned prior and the eigenvalues of $AA^T$. We observe that sample quality improves upon those in Figure~\ref{fig:sin:samples} and  the prior has learned a degenerate Gaussian with $\texttt{Indim}(\sP_Z)=1$, matching $\texttt{Indim}(\sP_d$). We also highlight that our model, in addition to learning the manifold support of the target distribution, can capture the `density' allocation on the manifold, see Appendix~\ref{app:toy} for further details.

\textbf{3D toy data (S-curve)}
We model the S-curve dataset  (Figure~\ref{fig:scurve:data}), where the data lies on a 2D manifold in a 3D space ($\texttt{Indim}(\sP_d){=}2$), see Appendix~\ref{app:network} for  details.  Our model learns a function $g$ to transform $\sP_d$ to $\sQ_Z$, where the latter lies on a 2D linear subspace in 3D space (see Figure \ref{fig:scurve:latent}). As discussed in Section~\ref{sec:identification}, we also conduct a linear dimension reduction 
to generate 2D representations, see Figure~\ref{fig:scurve:proj}. The colormap indicates correspondence between the data in 3D space and the 2D representation. Our method can be observed to successfully: (1) identify the intrinsic dimensionality of the data and (2) project the data into a 2D space that faithfully preserves the structure of the original distribution. In contrast, we find that a flow with fixed Gaussian prior fails to learn such a data distribution and generate meaningful representations, see Figure~\ref{fig:scurve:fix:x} and~\ref{fig:scurve:fix:z}.

\begin{figure*}[t]
\centering
\begin{subfigure}{0.23\linewidth}
\centering
\includegraphics[width=1\linewidth]{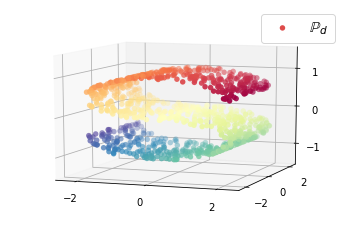} 
\caption{$\rvx\sim\sP_d$}
\label{fig:scurve:data}
\end{subfigure}
\begin{subfigure}{0.23\textwidth}
\centering
\includegraphics[width=1.0\linewidth]{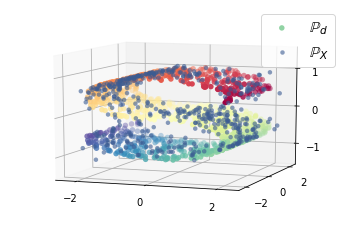} 
\caption{Our samples\label{fig:scurve:our:x}}
\end{subfigure}
\begin{subfigure}{0.23\linewidth}
\centering
\includegraphics[width=1\linewidth]{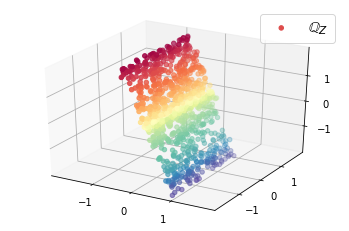}
\caption{Our representations}
\label{fig:scurve:latent}
\end{subfigure}
\begin{subfigure}{0.23\textwidth}
\centering
\includegraphics[width=1.0\linewidth]{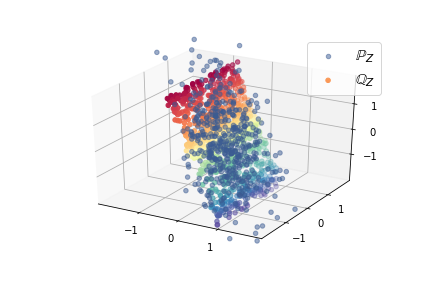}
\caption{Our Learned $\sP_z$ \label{fig:scurve:our:z}}
\end{subfigure}
\begin{subfigure}{0.23\linewidth}
\centering
\includegraphics[width=0.7\linewidth]{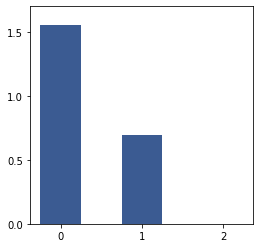}
\caption{Eigenvalues}
\label{fig:scurve:eigen}
\end{subfigure}
\begin{subfigure}{0.23\linewidth}
\centering
\includegraphics[width=1.0\linewidth]{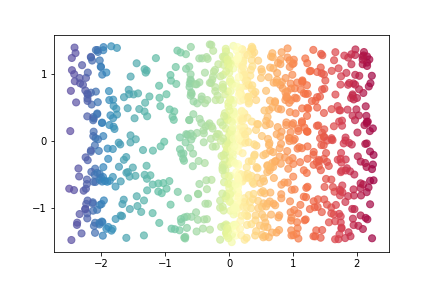}
\caption{$\rvz^{proj}=\rvz\mathbf{E}$\label{fig:scurve:proj}}
\end{subfigure}
\begin{subfigure}{0.23\textwidth}
\centering
\includegraphics[width=1.0\linewidth]{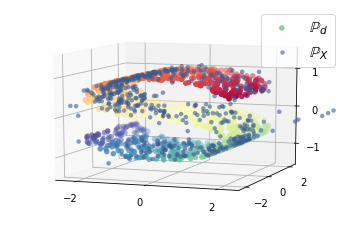} 
\caption{Trad. Flow samples\label{fig:scurve:fix:x}}
\end{subfigure}
\begin{subfigure}{0.23\textwidth}
\centering
\includegraphics[width=1.0\linewidth]{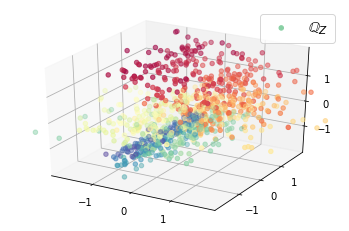} 
\caption{Trad. Flow representations\label{fig:scurve:fix:z}}
\end{subfigure}

\caption{(a) S-curve data  $\rvx\sim \sP_d$.\ (b) Samples generated from our learned model. (c) Latent representation $\rvz=g(\rvx)$, points are lying on a linear subspace. (d) Samples from the learned prior distribution. (e) Eigenvalues of the matrix $AA^T$, we deduce that $\texttt{Indim}(\sP_d)=2$.\ (f) Our representation after the dimensionality reduction $\rvz^{proj}=\rvz \mathbf{E}$. (g) and (h): Samples and  representations from a learned traditional flow model with a fixed Gaussian prior.\label{fig:scurve}} 
\end{figure*}

\begin{figure*}
\centering
\begin{subfigure}{0.22\linewidth}
\centering
\includegraphics[width=0.9\linewidth]{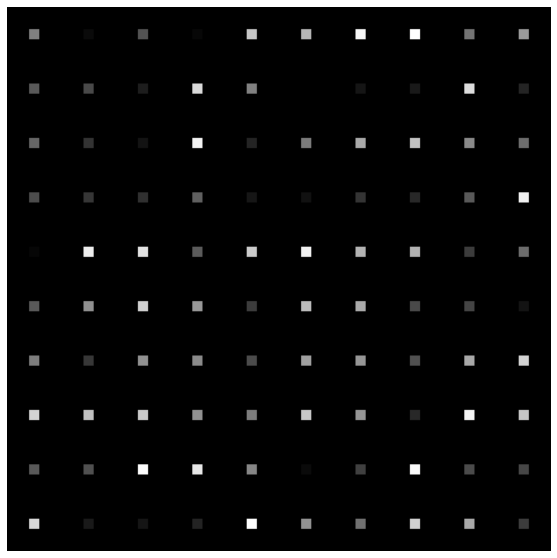} 
\caption{$\rvx\sim\sP_d$}
\label{fig:fading:squares:data}
\end{subfigure}
\begin{subfigure}{0.22\linewidth}
\centering
\includegraphics[width=0.9\linewidth]{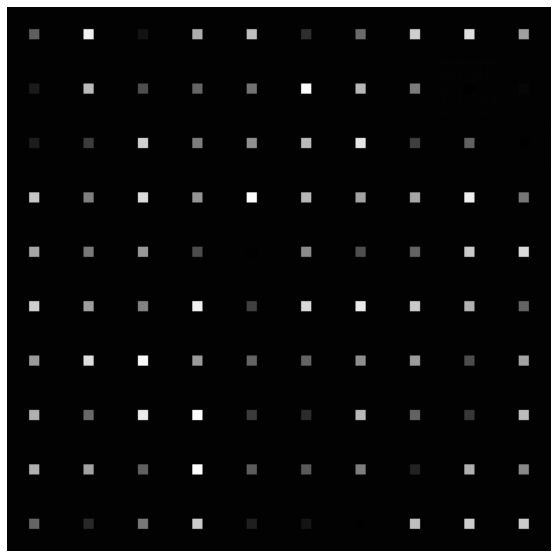}
\caption{Our model}
\label{fig:fading:squares:samples}
\end{subfigure}
\begin{subfigure}{0.22\linewidth}
\centering
\includegraphics[width=0.9\linewidth]{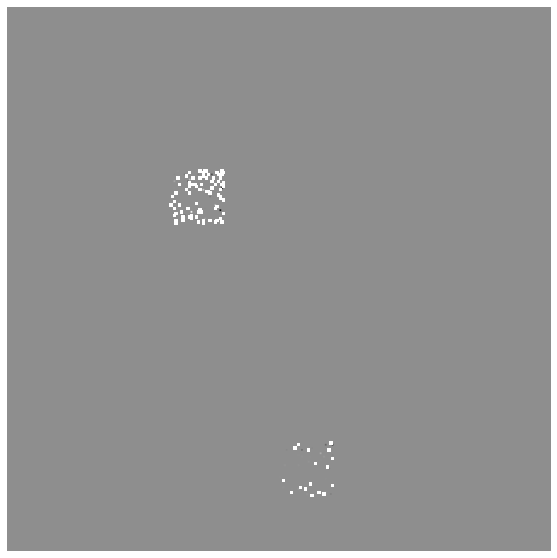}
\caption{Traditional flow}
\label{fig:fading:squares:traditional:flow}
\end{subfigure}
\begin{subfigure}{0.245\linewidth}
\centering
\includegraphics[width=1.1\linewidth]{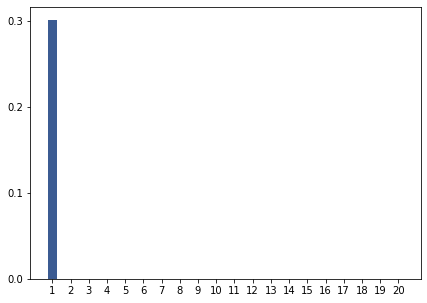}
\caption{Eigenvalues of $AA^T$}
\label{fig:fading:squares:eigen}
\end{subfigure}
\captionsetup[subfigure]{aboveskip=-12pt,belowskip=0pt}
\begin{subfigure}{\linewidth}
\centering
\includegraphics[width=1.0\linewidth]{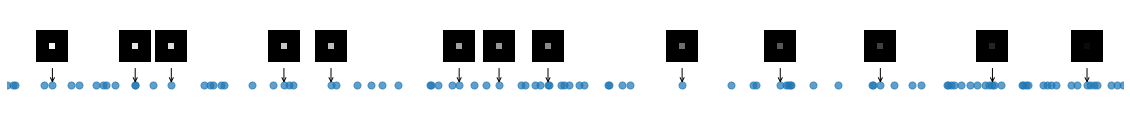}
\vspace{-3pt}
\caption{$\rvz^{proj}=\rvz\mathbf{E}$\label{fig:dot:representation}}
\end{subfigure}
\caption{(a) and (b) show samples from $\sP_d$ and our model, respectively. (c) shows  a traditional flow based model with a fixed Gaussian prior fails to fit the data distribution and cannot generate  valid samples. (d) the first 20 eigenvalues of the matrix $AA^T$. (e) shows the representation after applying dimensionality reduction. See text for further discussion.\label{fig:fading:square}}
\end{figure*}

\textbf{Fading Square dataset}
The \emph{fading square} dataset~\cite{rubenstein2018latent} was proposed in order to assess model behavior when data distribution and model possess differing intrinsic dimension and therefore affords a  relevant test of our work. The dataset consists of $32{\times}32$ images with $6{\times}6$ grey squares on a black background. The grey scale values are sampled from a uniform distribution with range $[0,1]$, so $\texttt{Indim}(\sP_d){=}1$. Figure~\ref{fig:fading:squares:data} shows samples from the dataset to which we fit our model and additional model details may be found in Appendix~\ref{app:exp}. Figure~\ref{fig:fading:squares:samples} shows samples from our trained model and Figure~\ref{fig:fading:squares:eigen} shows the first $20$ eigenvalues of $AA^T$ (ranked from high to low), we observe that only one eigenvalue is larger than zero and the others have converged to zero. This illustrates we have successfully identified the intrinsic dimension of $\sP_d$. We further carry out the dimensionality reduction process; the latent representation $\rvz$ is projected onto a 1D line and we plot the correspondence between the projected representation and the data in Figure \ref{fig:dot:representation}. Pixel values can be observed to decay as the 1D representation is traversed from left to right, indicating the representations are consistent with the properties of the original data distribution. In contrast, we find that the traditional flow model, with fixed 1024D Gaussian prior, fails to learn the data distribution; see Figure \ref{fig:fading:squares:traditional:flow}.

\textbf{Synthetic MNIST}
We further investigate model training using  digit images. It may be noted that, for real-world image datasets, the true intrinsic dimension is unknown. Therefore in order to further verify the correctness of our model's ability to identify intrinsic dimension, we first construct a synthetic dataset by fitting an implicit model $p_\theta(\rvx){=}\int \delta(\rvx-\nolinebreak g(\rvz)p(\rvz))d\rvz$ to the MNIST, and then sample from the trained model $\rvx\sim p_\theta(\rvx)$ in order to generate synthetic training data. The intrinsic dimension of the training data can then be 
approximated\footnote{For implicit model: 
$X=g(Z)$, the intrinsic dimension of the model distribution will be not-greater-than the dimension of the latent variable: $\texttt{Indim}(X)\leq \texttt{dim}(Z)$ \cite{arjovsky2017towards}. Further, if strictly $\texttt{Indim}(X)< \texttt{dim}(Z)$ this results in a `degenerated' case. When we fit the model on a data distribution where $\texttt{Indim}(X_d)\geq \texttt{dim}(Z)$, we assume that degeneration will not occur during training, resulting in $\texttt{Indim}(\sP_d)\approx\texttt{dim}(Z)$. For simplicity, we equate $\texttt{Indim}(\sP_d){=}\texttt{dim}(Z)$ in the main text.} 
via the dimension of the latent variable $Z$: 
$\texttt{Indim}(\sP_d){=}\texttt{dim}(Z)$. We construct two datasets with $\texttt{dim}(Z){=}5$ and $\texttt{dim}(Z){=}10$ such that $\texttt{Indim}(\sP_d){=}5$ and $\texttt{Indim}(\sP_d){=}10$, respectively. Since the generator of the implicit model is not constrained to be invertible, a diffeomorphism between data distribution and prior will not hold and will also not necessarily lie on a continuous manifold. We found when the continuous manifold assumption is not satisfied, training instabilities can occur. Towards alleviating this issue, we smooth the data manifold by adding small Gaussian noise (with standard deviation $\sigma_x{=}0.05$) to the training data. We note that adding Gaussian noise changes the intrinsic dimension of the training distribution towards alignment with the ambient dimension and therefore we mitigate this undesirable effect by annealing $\sigma_x$ after $2000k$ iterations with a factor of $0.9$ every $10k$ iteration. Experimentally, we cap a lower-bound Gaussian noise level of $0.01$ to help retain successful model training and find the effect of the small noise to be negligible when estimating the intrinsic dimension. In Figure~\ref{fig:mnist:eigen}, we plot the first twenty eigenvalues of $AA^T$ (ranked from high to low) after fitting two synthetic MNIST with $\texttt{dim}(\sP_d){=}\{5,10\}$. We can find that $5$ and $10$ eigenvalues are significantly larger than other values, respectively. The remaining non-zero eigenvalues can be attributed to the added small Gaussian noise. Therefore, our method can successfully model complex data distributions and identify the intrinsic dimension when the data distribution lies on a continuous manifold. We also provide model sample comparisons in Figure~\ref{fig:sample:mnist}, where we can find our model can achieve better sample quality compared to the traditional flow.
In the next section, we apply our model to real image data where this assumption may not be satisfied.

\subsection{Real World Data \label{sec:exp:real}}
\begin{figure*}[t]
\centering
\begin{subfigure}{0.3\textwidth}
    \centering
    \includegraphics[width=.75\linewidth]{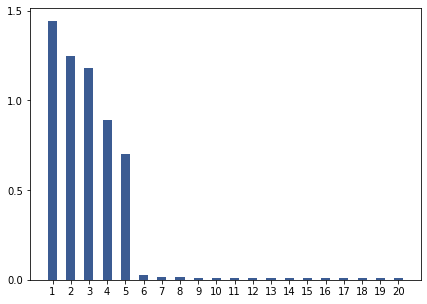}
    \caption{$\texttt{Indim}(\sP_d)=5$\label{fig:mnist:5}}
    \end{subfigure}
\begin{subfigure}{0.3\textwidth}
    \centering
    \includegraphics[width=.75\linewidth]{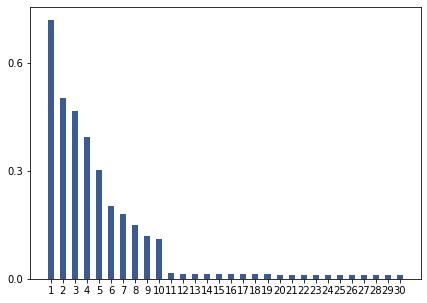}
    \caption{$\texttt{Indim}(\sP_d)=10$\label{fig:mnist:10}}
    \end{subfigure}
    \begin{subfigure}{0.3\textwidth}
    \centering
    \includegraphics[width=.75\linewidth]{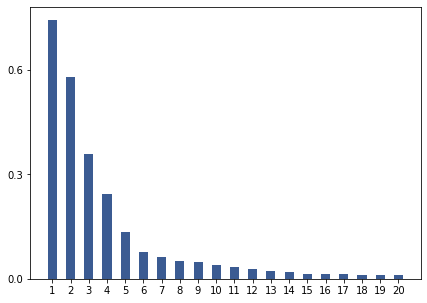}
    \caption{Real MNIST\label{fig:mnist:real}}
    \end{subfigure}
\caption{Eigenvalues estimated by our model trained on synthetic MNIST with $\texttt{Indim}(\sP_d)=\{5,10\}$ and real MNIST respectively.\label{fig:mnist:eigen}}
\end{figure*}

\begin{figure}[h]
\centering
\begin{subfigure}{0.45\linewidth}
    \centering
    \includegraphics[width=.9\linewidth]{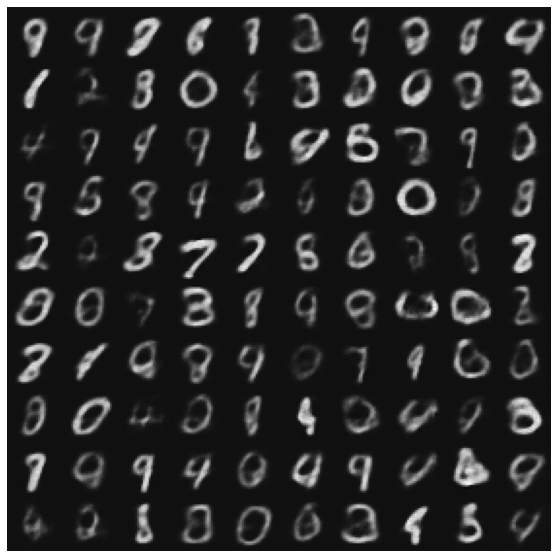}
    \caption{$\texttt{Indim}(\sP_d)=5$ (T)}
    \end{subfigure}
\begin{subfigure}{0.45\linewidth}
    \centering
    \includegraphics[width=0.9\linewidth]{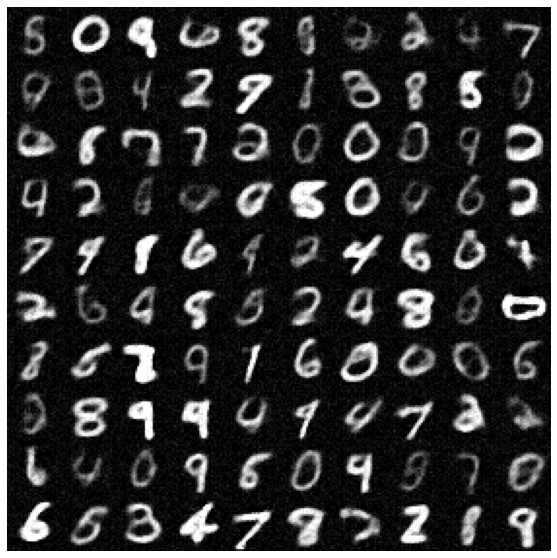}
    \caption{$\texttt{Indim}(\sP_d)=5$ (O)}
    \end{subfigure}

\begin{subfigure}{0.45\linewidth}
    \centering
    \includegraphics[width=.9\linewidth]{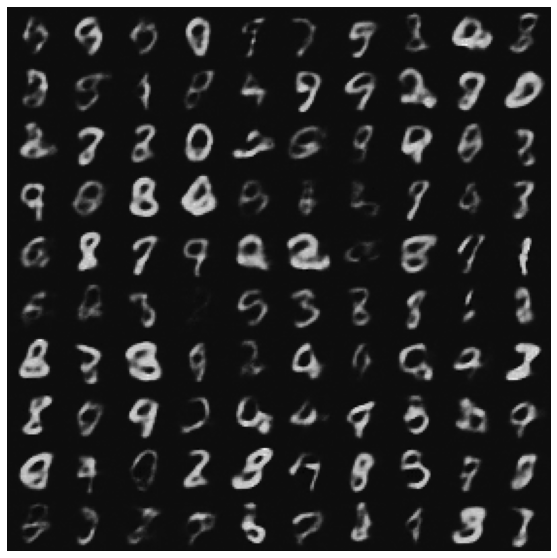}
   \caption{$\texttt{Indim}(\sP_d)=5$ (T)}    \end{subfigure}
\begin{subfigure}{0.45\linewidth}
    \centering
    \includegraphics[width=0.9\linewidth]{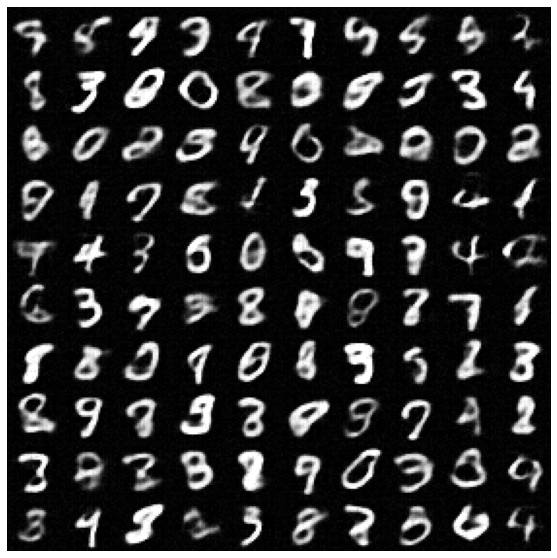}
   \caption{$\texttt{Indim}(\sP_d)=10$ (O)}   
    \end{subfigure}

\begin{subfigure}{0.45\linewidth}
    \centering
    \includegraphics[width=.9\linewidth]{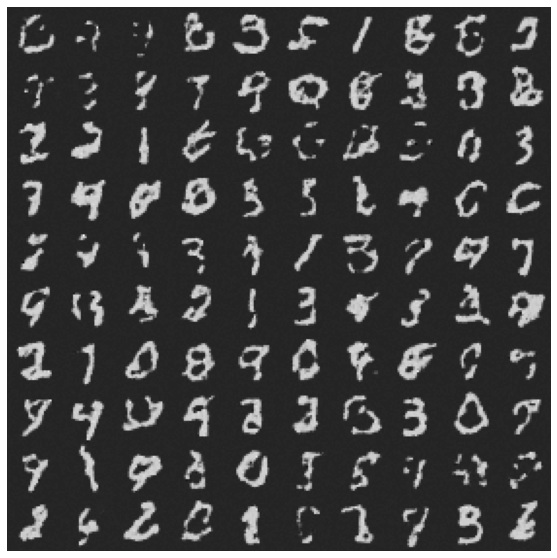}
    \caption{Real MNIST (T)}
    \end{subfigure}  
\begin{subfigure}{0.45\linewidth}
    \centering
    \includegraphics[width=0.9\linewidth]{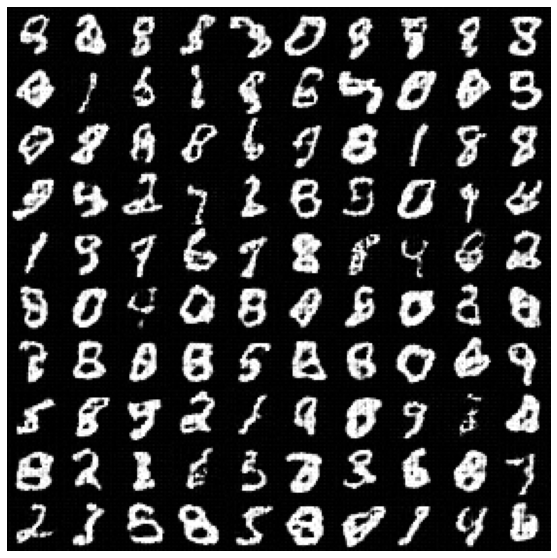}
    \caption{Real MNIST (O)}
    \end{subfigure}
\caption{We compare the  samples from a traditional non-volume preserving flow with a fixed prior (denoted as T) and our model 
(denoted as O) on synthetic MNIST with $\texttt{Indim}(\sP_d)=\{5,10\}$ and real MNIST, see Appendix~\ref{app:mnist} for experiment details. We find that our model can provide improved sample quality on these manifold datasets.
\label{fig:sample:mnist} }
\end{figure}

\textbf{Real MNIST} For the real MNIST, it was shown that digits have differing intrinsic dimensions~\cite{intrimnist3}. This suggests the MNIST may lie on \emph{several}, \emph{disconnected} manifolds with differing intrinsic dimensions. Although our model assumes that $\sP_d$ lies on one continuous manifold, it is interesting to investigate the case when this assumption is not fulfilled. We thus fit our model to the original MNIST and plot the eigenvalues in Figure~\ref{fig:mnist:eigen} (a) and (b). In contrast to Figures~\ref{fig:mnist:5}, \ref{fig:mnist:10}; 
the gap between eigenvalues predictably exhibits a less obvious step change. However, the values suggest the intrinsic dimension of MNIST lies between 11 and 14. This result is consistent with previous estimations, stating that the intrinsic dimension of MNIST is between 12 and 14~\cite{intrimnist1,intrimnist2,intrimnist3}, see also Figure~\ref{fig:sample:mnist} for the model samples comparisons.
Recent work~\cite{cornish2019relaxing} discusses fitting flow models to a $\sP_d$ that lies on disconnected components, by introducing a mixing prior. Such techniques may be easily combined with our method towards constructing more powerful flow models.

    
    



\begin{figure}
\vspace{-0.2cm}
\centering
\begin{subfigure}{0.48\linewidth}
\centering
\includegraphics[width=0.95\linewidth]{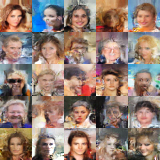}
\end{subfigure}
\begin{subfigure}{0.48\linewidth}
\centering
\includegraphics[width=0.95\linewidth]{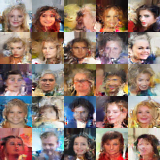}
\end{subfigure}
\caption{Samples from the proposed manifold prior with $K=2000$ (left) and $K=3000$ (right) respectively.\label{fig:color}}
\vspace{-0.3cm}
\end{figure}

\begin{figure}[h]
    \centering
\includegraphics[width=0.9\linewidth]{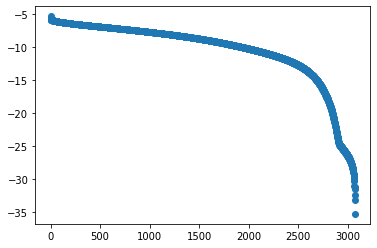}
    \caption{Eigen-values in log-space. We can see the eigenvalues has an exponential decay starting around 2700, which indicates that the intrinsic dimension of the CelebA (with size $32\times 32$) is around 2700.}
    \label{fig:log:eigen}
\end{figure}
\begin{figure}[h]
\centering
\begin{subfigure}{\linewidth}
\centering
\includegraphics[width=\linewidth]{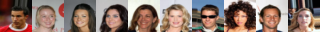}
\caption{True images}
\end{subfigure}
\begin{subfigure}{\linewidth}
\centering
\includegraphics[width=\linewidth]{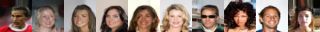}
\caption{$K=3000$}
\end{subfigure}
\begin{subfigure}{\linewidth}
\centering
\includegraphics[width=.95\linewidth]{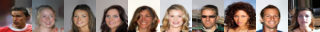}
\caption{$K=2800$}
\end{subfigure}
\begin{subfigure}{\linewidth}
\centering
\includegraphics[width=.95\linewidth]{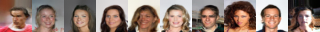}
\caption{$K=2700$}
\end{subfigure}
\begin{subfigure}{\linewidth}
\centering
\includegraphics[width=.95\linewidth]{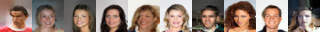}
\caption{$K=2600$}
\end{subfigure}
\begin{subfigure}{\linewidth}
\centering
\includegraphics[width=.95\linewidth]{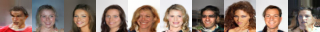}
\caption{$K=2500$}
\end{subfigure}
\caption{Reconstructions with different dimensional  representations. We can see the quality of the reconstruction slightly decreases when $K$ is lower than 2700, which empirically verified that the intrinsic dimension of the data distribution is around 2700.}
\label{fid:rec}
\end{figure}

\textbf{CelebA}
To model CelebA, we first
centre-crop and resize each image  to $32{\times}32{\times}3$. Pixels take discrete values from $[0,\ldots,255]$ and we follow a standard dequantization process~\cite{uria2013rnade}: $x{=}(x+\text{Uniform}(0,1))/256$ to transform each pixel to a continuous value in $[0,1]$. Similar to the MNIST experiment, we add small Gaussian noise (with standard deviation $\sigma_x{=}0.01$) for twenty initial training epochs in order to smooth the data manifold and then anneal the noise with a factor of $0.9$ for a further twenty epochs. Additional training and network structure details can be found in Appendix~\ref{app:network:color}. 

We aim to empirically verify if the learned model is concentrated on a low-dimensional manifold and examine the validity of the estimated intrinsic dimensions. Following~\cite{ross2021tractable}, we study if the data can be reconstructed from a low-dimensional linear manifold specified by $AA^T$. In Figure~\ref{fig:log:eigen}, we plot the learned eigenvalues  $\Lambda$ in log space and observe that eigenvalues have an exponential decay starting at ${\sim}2700$, which suggests that the intrinsic dimension of the CelebA is around 2700. For reconstructions, we  first calculate the low dimensional representation of image $\mathbf{x}$ by transforming it to the latent space with the inverse flow $\mathbf{z}=g(\mathbf{x})$ ($\mathbf{z}\in\mathbb{R}^D$) and then conduct a linear projection $\mathbf{z}^{proj}=\mathbf{z}\mathbf{E}_k$ to obtain the representation $\mathbf{z}^{proj}\in\mathbb{R}^K$, where $\mathbf{E}_k$ contains the first $K$ eigenvectors ranked by the corresponding eigenvalues (see Section~\ref{sec:identification}). To reconstruct $\mathbf{x}$ from $\mathbf{z}^{proj}$, we firstly conduct an inverse projection $\hat{\mathbf{z}}=\mathbf{z}^{proj}\mathbf{E}_k^T$ and then obtain the reconstruction $\hat{\mathbf{x}}$ by letting $\hat{\mathbf{x}}=f(\hat{\mathbf{z}})$, where $f=g^{-1}$. In Figure \ref{fid:rec} we show reconstructions with differing $K$ and may observe that good reconstruction can be obtained when $K$ is greater than $2700$. The quality goes down as $K$ decreases, which empirically verifies the intrinsic dimension of the data distribution to be around ${\sim}2700$, which is consistent with our model estimation (as shown in Figure~\ref{fig:log:eigen}).

We also compare the sample quality of our model with the recently proposed CEF model~\cite{ross2021tractable}, which proposes to learn a flow on manifold distributions. Distinct from our model, where different $K$ values can be selected post training, the CEF work requires an intrinsic model dimension 
to be pre-specified. We follow~\cite{ross2021tractable}; select $K=512$ and train the model on the CelebA ($32{\times}32$) dataset. Our model achieves better FID compared to CEF (see Table~\ref{tab:fid} in the Appendix for FID comparisons). The visualizations of the samples can be found in Figure~\ref{fig:celebe:sample} (our model with $K=\{2000,3000\}$) and Appendix~\ref{app:comparison} (comparisons of both models). We also train our model on SVHN~\cite{svhn} and CIFAR10~\cite{cifar10}
, see Appendix~\ref{app:exp} for a detailed discussion.

\section{Related Work}

Classic latent variable generative models assume that data distributions lie \emph{around} a low-dimensional manifold, for example, the Variational Auto-Encoder~\cite{kingma2013auto} and the recently introduced Relaxed Injective Flow~\cite{kumar2020regularized} (which uses an $L2$ reconstruction loss in the objective, implicitly assuming that the model has a Gaussian observation distribution, with isotropic variance). Such methods typically assume that observational noise is not degenerated (\eg a fixed Gaussian distribution), therefore the model distribution is absolutely continuous and maximum likelihood learning is thus well defined~\cite{loaiza2022diagnosing,spread}. However, common distributions such as natural images usually do not have Gaussian observational noise~\cite{zhao2017towards}. Therefore, we focus on modelling distributions that lie \emph{on} a low-dimensional manifold.

The study of manifold learning for nonlinear dimensionality reduction \cite{cayton2005algorithms} and intrinsic dimension estimation \cite{camastra2016intrinsic} is a rich field with an  extensive set of tools. However, most methods commonly do not model the data density on the manifold and are thus not used for the same purpose as the models introduced in our work. 
There are however a number of recent works that introduce normalizing flows on manifolds that we now highlight and relate to our approach. 





Several works define flows on manifolds with prescribed charts. \cite{gemici2016normalizing} generalized flows from Euclidean spaces to Riemannian manifolds by proposing to 
map points from the manifold $\mathcal{M}$ to $\mathbb{R}^K$, apply a normalizing flow in this space and then mapping back to $\mathcal{M}$. The technique has since been further extended to Tori and Spheres~\cite{rezende2020normalizing}. In contrast to our work, these methods require knowledge of the intrinsic dimension $K$ and a parameterization of the coordinate
chart of the data manifold. 
Several recently proposed manifold flow models can learn the data distribution without providing a chart mapping a priori. M-flow~\cite{brehmer2020flows} proposed an algorithm  that maps the data distribution to a lower-dimensional space with an auto-encoder and uses a flow in the lower-dimensional space to learn data distributions. Rectangular flow~\cite{caterini2021rectangular} and Conformal Embedding Flows~\cite{ross2021tractable} propose to directly build injective  mappings from low-dimension space to high-dimension space.
However, their methods still require that the dimensionality of the manifold is known. In~\cite{brehmer2020flows}, it is proposed that dimensionality can be learned either by a brute-force solution or through a trainable variance in the density function. The brute-force solution is clearly infeasible for data embedded in extremely high dimensional space, as is often encountered in deep learning tasks. Use of a trainable variance is natural and similar to our approach. However, as discussed at the beginning of this paper, without carefully handling the KL or MLE term in the objective, a vanishing variance parameter will result in the wild behaviour of the optimization process since these terms are not well defined.


The GIN model considered in~\cite{sorrenson2020disentanglement} could recover the low-dimensional generating latent variables by following their identifiability theorem. However, the assumptions therein require knowledge of an auxiliary variable, \ie the label, which is not required in our model. Behind this difference is the essential discrepancy between the concept of informative dimensions and intrinsic dimensions. The GIN model discovers the latent variables that are informative in a given context, defined by the auxiliary variable $u$ instead of the true intrinsic dimensions. In their synthetic example, the ten-dimensional data is a nonlinear transformation of ten-dimensional latent variables where two out of ten are correlated with the labels of the data and the other eight are not. In this example, there are two informative dimensions, but ten intrinsic dimensions. Nevertheless, our method for intrinsic dimension discovery can be used together with informative dimension discovery methods to discover finer structures of data. 

In recent work~\cite{tempczyk2022lidl}, the authors made an intriguing observation regarding the impact of adding Gaussian noise $\mathcal{N}(0,\sigma^2I)$ to a given data distribution. Specifically, they found that the change in log-likelihood, resulting from adding noise with different values of $\sigma$ is approximately linear in the logarithm of $\sigma$, with a proportionality constant that corresponds to the difference between the intrinsic and ambient dimensions of the data. Although both methods involve fitting flows to a noised data distribution, our method for estimating the intrinsic dimension is different: we estimate the intrinsic dimension using the rank of a learned degenerate Gaussian prior. We consider the future study of connections between these approaches may help to afford a 
deeper understanding of intrinsic dimensionalities with respect to complex data distributions.

\section{Limitations and Future Work}
In our work, we present a novel manifold flow model that utilizes a learnable generalized Gaussian prior to capture the low-dimensional manifold data distribution and identify their intrinsic dimensions. We demonstrated the
benefits of our model in terms of sample generation and representation quality.
While our model offers a promising step forward on the modelling of manifold distributions, it is important to note that we rely on a somewhat strong assumption: that the data distribution lies on a continuous manifold. This assumption may not hold for real-world image data, which typically exhibit complex and varied distributions. 
We conjecture that this limitation can be partially alleviated through the incorporation of 
prior techniques developed by~\cite{cornish2019relaxing}; affording relaxation of the continuous manifold assumption and thus enable the flow model to capture data distributions that lie on a set of disconnected manifolds. We leave this to future work.


\clearpage
\newpage
\bibliography{ref}

\appendix
\onecolumn
\section{Maximum Likelihood Estimation and KL divergence\label{app:mle:kl}}
Given data $x_1,x_2,\ldots,x_N$ sampled independently from the true data distribution $\sP_d$, with density function $p_d(x)$, we want to fit the model density $p(x)$\footnote{For brevity, we use notation $p(x)$ to represent the model $p_X(x)$ unless otherwise specified.} to the data. A popular choice to achieve this involves minimization of the KL divergence where:
\begin{align}
    &\KL(p_d(x)||p(x))=\int p_d(x)\log p_d(x)dx-\int p_d(x)\log p(x)dx\\
    &=-\int p_d(x)\left(\log p_Z\left(g(x)\right)+\log \left|\det\left(\frac{\partial g}{\partial x}\right)\right|\right)dx+const..
\end{align}
Since we can only access samples from $p_d(x)$, we approximate the integral by Monte Carlo sampling
\begin{equation}
    \KL(p_d(x)||p(x))\approx-\frac{1}{N}\sum_{n=1}^N\log p(x_n)+const..
\end{equation}
Therefore, minimizing the KL divergence between the data distribution and the model is (approximately) equivalent to Maximum Likelihood Estimation (MLE).

When $p(x)$ is a flow-based model with invertible flow function $f:Z\rightarrow X, g=f^{-1}$, minimizing the KL divergence in $X$ space is equivalent to minimizing the KL divergence in the $Z$ space. We let $X_d$ be the random variable of the data distribution and define $Z_\sQ: Z_\sQ=g(X_d)$ with density $q(z)$, so $q(z)$ can be represented as 
\begin{align}
        q(z)&=\int \delta(z-g(x))p_d(x)dx.
\end{align}
Let $p(z)$ be the density of the prior distribution $\sP_Z$, the KL divergence in $Z$ space can be written as
\begin{align}
    \KL(q(z)||p(z))&=\underbrace{\int q(z)\log q(z)dz}_{\text{Term 1}}-\underbrace{\int q(z)\log p(z)dz}_{\text{Term 2}}.
\end{align}
Term 1: using the properties of transformation of random variables~\citet[pp. 660]{papoulis2002probability}, the negative entropy can be written as 
\begin{align}\label{eq:app:ne}
\int q(z)\log q(z)dz
     =\underbrace{\int p_d(x)\log p_d(x)dx}_{const.}-\int p_d(x) \log\left|\det\left(\frac{\partial g}{\partial x}\right)\right|dx.
\end{align}
Term2: the cross entropy can be written as
\begin{align}\label{eq:app:ce}
    \int q(z)\log p(z)dz&=\int\int \delta(z-g(x))p_d(x)\log p(z)dzdx\\
    &=\int p_d(x) \log p(g(x))dx.
\end{align}
Therefore, the KL divergence in $Z$ space is equivalent to the KL divergence in $X$ space
\begin{align}
    \KL(q(z)||p(z))=\KL(p_d(x)||p(x)).
\end{align}
We thus build the connection between MLE and minimizing the KL divergence in $Z$ space.
\section{Entropy}
\subsection{An example}\label{app:an:example}
Assume a 2D Gaussian random variable $X$ with covariance $\begin{bmatrix}
1 & 0 \\
0 & 1
\end{bmatrix}$. There exits two volume-perserving flows  $g$ with parameters $\theta_1$ and $\theta_2$ ($\theta_1\neq \theta_2$) to make  $Z_1=g_{\theta_1}(X)$ and $Z_2=g_{\theta_2}(X)$ be Gaussians with covariance $\begin{bmatrix}
2 & 0 \\
0 & \frac{1}{2}
\end{bmatrix}$  and $\begin{bmatrix}
3 & 0 \\
0 & \frac{1}{3}
\end{bmatrix}$ respectively. In this case,  the entropy $\H(Z_1)=\H(Z_2)=\H(X)$ and does not depend on $\theta$. However, for a given Gaussian variable $K$,  $\H(Z_1+K)\neq \H(Z_2+K)$ and $\H(g_\theta(X)+K)$ will depend on $\theta$.  For example, the distribution of $K$ is a Gaussian with covariance $\begin{bmatrix}
1 & 0 \\
0 & 1
\end{bmatrix}$, so $Z_1+K$ is a Gaussian with covariance $\begin{bmatrix}
3 & 0 \\
0 & \frac{3}{2}
\end{bmatrix}$ and $Z_2+K$ is a Gaussian with covariance $\begin{bmatrix}
4 & 0 \\
0 & \frac{4}{3}
\end{bmatrix}$, so $\H(Z_1+K)\neq \H(Z_2+K)$. A similar example can be constructed when $X$ is not absolutely continuous.

\subsection{Z is an absolutely continuous random variable\label{app:ac:entropy}}
For two mutually independent absolutely continuous random variable $Z$ and $K$, the mutual information between $Z+K$ and $K$ is
\begin{align}
   \I(Z+K,K)&=\H(Z+K)-\H(Z+K|K)\\&=\H(Z+K)-\H(Z|K)\\&=\H(Z+K)-\H(Z).  
\end{align}
  
The last equality holds because $Z$ and $K$ are independent. Since mutual information $\I(Z+K,K)\geq 0$, we have 
\begin{equation}
    \H(Z)\leq \H(Z+K) =\H(Z)+\I(Z+K,K).
\end{equation}

Assume $K$ has a Gaussian distribution with 0 mean and variance $\sigma_Z^2$\footnote{The extension to higher dimensions is straightforward.}. When $\sigma_Z^2=0$, $K$ degenerates to a delta function, so $Z+K=Z$ and  
\begin{equation}
    \I(Z+K,K)=\H(Z+K)-\H(Z)=0,
\end{equation}
this is because the mutual information  between an a.c. random variable and a singular random variable is still well defined, see \citet[Theorem 10.33]{yeung2008information}.
Assume $K_1$, $K_2$ are Gaussian random variables with 0 mean and variances $\sigma_1^2$ and $\sigma_2^2$ respectively. Without loss of generality, we assume $\sigma_1^2>\sigma_2^2$ and $\sigma_1^2=\sigma_2^2+\sigma_\delta^2$, and let $K_\delta$ be the random variable of a Gaussian that has 0 mean and variance $\sigma_\delta^2$ such that $K_1=K_2+K_\delta$. By the data-processing inequality, we have
\begin{align}
\I(Z+K_2,K_2)\leq \I(Z+K_2+K_\delta,K_2+K_\delta)=\I(Z+K_1,K_1).
\end{align}
Therefore, $\I(Z+K,K)$ is a monotonically decreasing function when $\sigma_Z^2$ decreases and when $\sigma_Z^2\rightarrow 0$, $I(Z+K,K) \rightarrow 0$.

\subsection{Upper bound of the spread KL divergence}\label{app:upper:bound}
In this section, we show that leaving out the entropy term $\H(Z_{\tilde{\sQ}})$ in \eqref{eq:spread:kl} is equivalent to minimizing an \emph{upper bound} of the spread KL divergence.

For singular random variable $Z_{\sQ}=g(X_d)$ and absolutely continuous random variable $K$ that are independent, we have 
\begin{align}
    \H(Z_{\sQ}+K)-\H(K)&=\H(Z_{\sQ}+K)-\H(Z_{\sQ}+K|Z_{\sQ})\\&=\I(Z_{\sQ}+K,Z_{\sQ})\geq 0.
\end{align}
The second equation is from the definition of Mutual Information (MI); the MI between an a.c. random variable and a singular random variable is well defined and always positive, see \citet[Theorem 10.33]{yeung2008information} for a proof.

Therefore, we can construct an upper bound of the spread KL objective in \eqref{eq:spread:kl}
\begin{align}
    \KL(\tilde{q}||\tilde{p})&=\underbrace{\int q(\tilde{\rvz})\log  q(\tilde{\rvz})d\tilde{\rvz }}_{-\H(Z_\sQ+K)}-\int q(\tilde{\rvz})\log p(\tilde{\rvz})d\tilde{\rvz}\\
    &=\underbrace{-\H(K)}_{const.}-\underbrace{\I(Z_{\sQ}+K,Z_{\sQ})}_{\geq 0} -\int q(\tilde{\rvz})\log p(\tilde{\rvz})d\tilde{\rvz}.
    \\
    &\leq \underbrace{-\H(K)}_{const.} -\int q(\tilde{\rvz})\log p(\tilde{\rvz})d\tilde{\rvz}.
\end{align}
Therefore, ignoring the negative entropy term during training is equivalent to minimizing an upper bound of the spread KL objective, and the gap between the bound and the true objective is $\I(Z_{\sQ}+K,Z_{\sQ})$. Unlike the case when $Z_{\sQ}$ is a.c., $\I(Z_{\sQ}+K,Z_{\sQ})$ will be close to $0$ when the variance of $K$ is sufficiently small, we do not have the same result when $Z_{\sQ}$ is not absolutely continuous. However, we show that, under some assumptions of the flow function, the gap  $\I(Z_{\sQ}+K,Z_{\sQ})$ can be upper bounded.

\subsection{Empirical evidence for ignoring the negative entropy \label{app:empirical:evidence}}

In this section, we first introduce the approximation technique to compute the negative entropy term, and then discuss the contribution of this term during training. The negative entropy of random variable $Z_{\tilde{Q}}$ is
\begin{align}\label{eq:neg_ent}
    -\H(Z_{\tilde{\sQ}})=\int  q(\tilde{\rvz})\log  q(\tilde{\rvz})d\tilde{\rvz},
\end{align}
where 
\begin{align}
     q(\tilde{\rvz})=\int_{\rvz} p_K(\tilde{\rvz}-\rvz)d\sQ_Z,
\end{align}
and $p_K$ is the density of a Gaussian with diagonal covariance $\sigma^2_ZI$. We first approximate $\tilde{q}(\tilde{\rvz})$ by a mixture of Gaussians
\begin{align}
    q(\tilde{\rvz})\approx \frac{1}{N}\sum_{n=1}^N \mathcal{N}(\tilde{\rvz}; \rvz^n,\sigma^2_ZI)\equiv \hat{q}^N(\tilde{\rvz})
\end{align}
where $\rvz^n$ is the $n$th sample from distribution $\sQ_z$ by first sampling $\rvx^n\sim \sP_d$ and letting $\rvz^n=g(\rvx^n)$. We denote the random variable of this Gaussian mixture as $\hat{Z}^N_{\tilde{\sQ}}$ and approximate
\begin{align}
    -\H(Z_{\tilde{\sQ}})\approx -\H(\hat{Z}^N_{\tilde{\sQ}}).
\end{align}
However, the entropy of a Gaussian mixture distribution does not have a closed form, so we further conduct a first order Taylor expansion approximation \cite{huber2008entropy}
\begin{align}
    -\H(\hat{Z}^N_{\tilde{\sQ}})\approx \frac{1}{N}\sum_{n=1}^N \log \hat{q}^N(\tilde{\rvz}=\rvz^n),
\end{align}
this approximation is accurate for small $\sigma_Z^2$. Finally we have our approximation
\begin{align}\label{eq:ent_approx}
    -\H(Z_{\tilde{\sQ}})\approx \frac{1}{N}\sum_{n=1}^N \log \hat{q}^N(\tilde{\rvz}=\rvz^n).
\end{align}
To evaluate the contribution of the negative entropy, we train our flow model on both low dimensional data (Toy datasets: 2D) and high dimensional data (Fading square dataset: 1024D) by optimization that uses two objectives: (1) ignoring the negative entropy term in \eqref{eq:spread_KL_divergence} and (2) approximating the negative entropy term in \eqref{eq:spread_KL_divergence} using the approximation discussed above. During training, we keep track of the value of the entropy $\H(Z_{\tilde{\sQ}})$ (using the approximation value) in both objectives. We let $N$ equal the batch size when approximating the entropy. Additional training details remain consistent with those described in Appendix~\ref{app:exp}.

In Figure \ref{fig:entropy_compar}, we plot the (approximated) entropy value $\H(Z_{\tilde{\sQ}})$ during training for both experiments. We find the difference between having the (approximated) negative entropy term and ignoring the negative entropy to be negligible. We leave rigorous theoretical investigation on the effects of discarding the entropy term during training to future work.

\begin{figure}[h]
\centering
\begin{subfigure}{0.45\textwidth}
\centering
\includegraphics[width=1.0\linewidth]{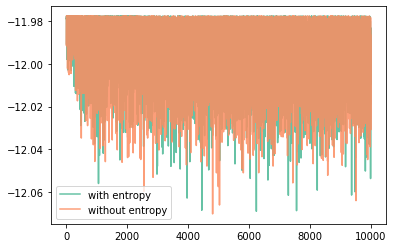} 
\caption{Toy dataset}
\end{subfigure}
\begin{subfigure}{0.45\textwidth}
\centering
\includegraphics[width=1.0\linewidth]{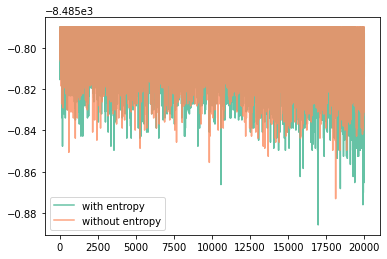} 
\caption{Fading square dataset}
\end{subfigure}
\caption{Figure (a) and (b) show the (approximated) entropy value using two different training objectives, for two different experiments.\label{fig:entropy_compar}}
\end{figure}

\section{Experiments\label{app:exp}}
We conduct all our experiments on one single NVDIA GeForce RTX 2080 Ti GPU.
\subsection{Network architecture for toy and MNIST dataset\label{app:network}}
The flow network we use consists of incompressible affine coupling layers \cite{sorrenson2020disentanglement,dinh2016density}.  Each coupling layer splits a $D$-dimensional input $\rvx$ into two parts $\rvx_{1:d}$ and $\rvx_{d+1:D}$. The output of the coupling layer is 
\begin{align}
    \rvy_{1:d}&=\rvx_{1:d}\\
    \rvy_{d+1:D}&=\rvx_{d+1:D}\odot \exp(s(\rvx_{1:d}))+t(\rvx_{1:d}),
\end{align}
where $s: \mathbb{R}^{d}\rightarrow \mathbb{R}^{D-d}$ and $t: \mathbb{R}^{d}\rightarrow \mathbb{R}^{D-d}$ are scale and translation functions parameterized by neural networks, $\odot$ is the element-wise product. The log-determinant of the Jacobian of a coupling layer is the sum of the scaling function $\sum_j s(\rvx_{1:d})_j$. To make the coupling transform volume preserving, we normalize the output of the scale function, so the $i$-th dimension of the output is
\begin{equation}
    \tilde{s}(\rvx_{1:d})_i=s(\rvx_{1:d})_i-\frac{1}{D-d}\sum_{j}s(\rvx_{1:d})_j,
\end{equation}
and the log-determinant of the Jacobian is $\sum_i \tilde{s}(\rvx_{1:d})_i=0$. We compare a volume-preserving flow with a learnable prior (normalized $s(\cdot)$) to a non-volume preserving flow with a fixed prior (unnormalized $s(\cdot)$). In this way both models have the ability to adapt their `volume' to fit the target distribution, retaining comparison fairness.

In our affine coupling layer, the scale function $s$ and the translation function $t$ have two types of structure: fully connected net and convolution net. Each fully connected network is a 4-layer neural network with hidden-size 24 and Leaky ReLU with slope 0.2. Each convolution net is a 3-layer convolutional neural network with hidden channel size 16, kernel size $3{\times}3$ and padding size 1. The activation is Leaky ReLU with a slope 0.2. 
The downsampling decreases the image width and height by a factor of 2 and increases the number of channels by 4 in a checkerboard-like fashion~\cite{sorrenson2020disentanglement,jacobsen2018revnet}. When multiple convolutional nets are connected together, we randomly permute the channels of the output of each network except the final one. In Table \ref{tb:structure:toy}, \ref{tb:structure:scurve},
\ref{tb:structure:sauqre},
\ref{tb:structure:mnist}, we show the network structures for our four main paper experiments.

\subsection{Toy dataset\label{app:toy}}
\begin{figure}[h]
\centering
\begin{subfigure}{0.3\textwidth}
\centering
\includegraphics[width=1.0\linewidth]{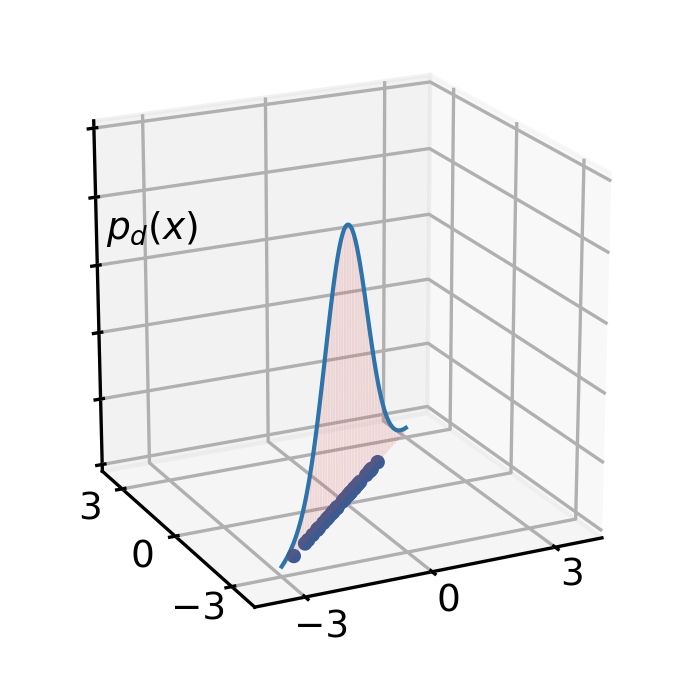}
\caption{Data distribution}
\label{fig:linedata}
\end{subfigure}
\begin{subfigure}{0.3\textwidth}
\centering
\includegraphics[width=1.0\linewidth]{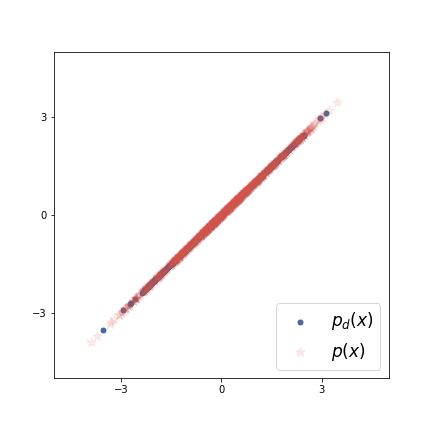}
\caption{Fixed Gaussian prior: $X$ space}
\end{subfigure}
\begin{subfigure}{0.3\textwidth}
\centering
\includegraphics[width=1.0\linewidth]{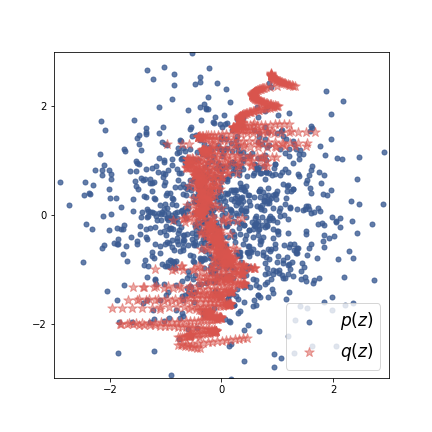}
\caption{Fixed Gaussian prior: $Z$ space}
\label{fig:line:fix:latent}
\end{subfigure}

\begin{subfigure}{0.3\textwidth}
\centering
\includegraphics[width=1.0\linewidth]{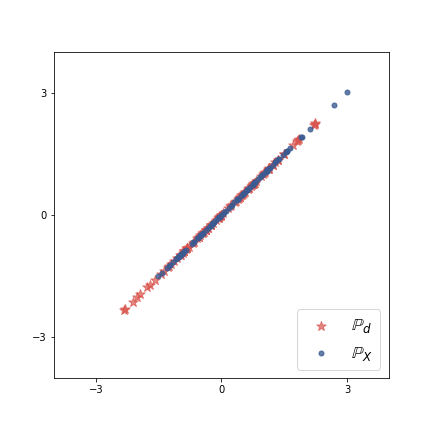}
\caption{Our method: $X$ space}
\end{subfigure}
\begin{subfigure}{0.3\textwidth}
\centering
\includegraphics[width=1.0\linewidth]{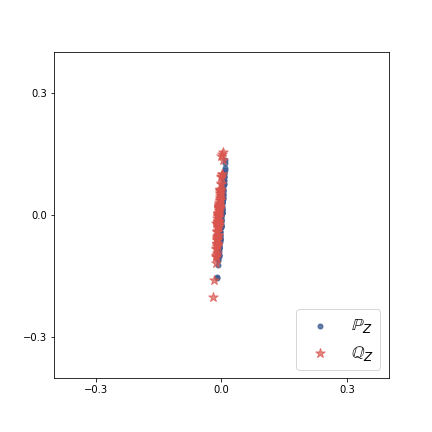}
\caption{Our method: $Z$ space}
\label{fig:line:latent}
\end{subfigure}
\begin{subfigure}{0.3\textwidth}
\centering
\includegraphics[width=0.6\linewidth]{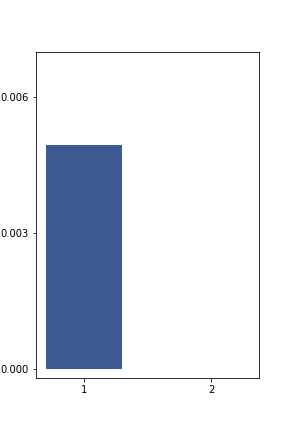}
\caption{Eigenvalues of $AA^T$}
\label{fig:line:eigen}
\end{subfigure}

\caption{(a) shows the samples from the data distribution. (b) and (d) show samples from a flow with a fixed Gaussian prior and our method, (c) and (e) show the latent space in both models. In (f), we plot the eigenvalues of $AA^T$.\label{fig:line}}
\end{figure}

We also construct a second dataset and train a flow model using the same training procedure discussed in Section~\ref{sec:exp}. Figure~\ref{fig:linedata} shows the samples from the data distribution $\sP_d$. Each data point is a 2D vector $\rvx=[x_1,x_2]$ where $x_1\sim \mathcal{N}(0,1)$ and $x_2=x_1$, so $\texttt{Indim}(\sP_d)=1$.
Figure \ref{fig:line:eigen} shows that the prior $\sP_Z$ has learned the true intrinsic dimension $\texttt{Indim}(\sP_Z)=\texttt{Indim}(\sP_d)=1$. We train our model using  learning rate $3\times 10^{-4}$ and batch size 100 for $10k$ iterations. We compare to samples drawn from a flow model that uses a fixed 2D Gaussian prior, with results shown in Figure~\ref{fig:line}. We can observe, for this simple dataset, flow with a fix Gaussian 
prior can generate reasonable samples, but the `curling up' behavior, discussed in the main paper, remains highly evident in the $Z$ space, see Figure~\ref{fig:line:fix:latent}.

\begin{figure}[h]
\centering
\begin{subfigure}{0.4\textwidth}
\centering
\includegraphics[width=0.8\linewidth]{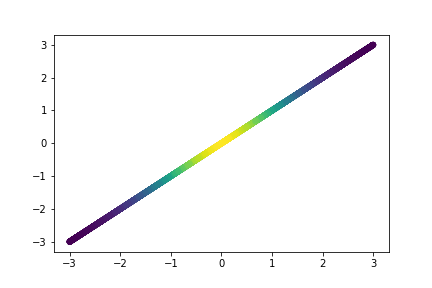} 
\caption{Ground truth}
\end{subfigure}
\begin{subfigure}{0.4\textwidth}
\centering
\includegraphics[width=0.8\linewidth]{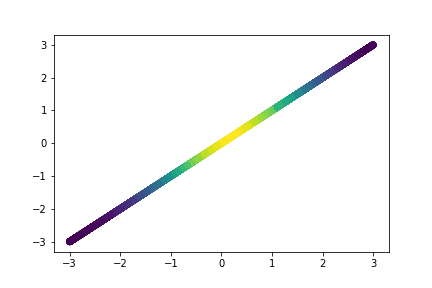} 
\caption{Our method}
\end{subfigure}
\begin{subfigure}{0.4\textwidth}
\centering
\includegraphics[width=0.8\linewidth]{img/sin_true_density.png}
\caption{Ground truth}
\end{subfigure}
\begin{subfigure}{0.4\textwidth}
\centering
\includegraphics[width=0.8\linewidth]{img/sin_our_density.png} 
\caption{Our method}
\end{subfigure}
\caption{(a) and (c) show the ground truth `density allocation' on the manifold for two toy datasets, (b) and (d) show the `density allocation' learned by our models. \label{fig:manifold:density}}
\end{figure}

\textbf{Manifold density} We also plot the `density allocation' on the manifold for the two toy datasets. For example, for the data generation process $\rvx=[x_1,x_2]$ where $x_1\sim p=\mathcal{N}(0,1)$ and $x_2=x_1$, we use the density value $p(x=x_1)$ to indicate the `density allocation' on the 1D manifold. To plot the `density allocation' of our learned model, we first sample $\rvx_s$ uniformly from the support of the data distribution, the subscript `s' here means that they only contain the information of the \emph{support}. Specifically, since $\rvx_s=[x^s_1,x^s_2]$, we sample $x^s_1\sim p=\mathcal{U}(-3\sigma,3\sigma)$ ($\sigma$ is the standard deviation of $\mathcal{N}(0,1)$, $\mathcal{U}$ is the uniform distribution) and let $x^s_2=x^s_1$ or $x^s_2=\sin(x^s_1)$, depending on which dataset is used. We use the 
projection procedure that was described in Section~\ref{sec:exp} to obtain the projected samples $\rvz_s^{proj}$, so $\rvz_s^{proj}\in\sR^{\texttt{Indim}(\sP_d)}$. We also project the learned prior $\sP_Z$ to $\sR^{\texttt{Indim}(\sP_d)}$ by constructing $\sP^{proj}_Z$ as a Gaussian with zero mean and a diagonal covariance contains the non-zeros eigenvalues of $AA^T$. Therefore $\sP^{proj}_Z$ is a.c. in $\sR^{\texttt{Indim}(\sP_d)}$ and we denote its density function as $p^{proj}(\rvz)$. We then use the density value $p^{proj}(\rvz=\rvz_s^{proj})$ to indicate the `density allocation' at the location of $\rvx_s$ on the manifold support. In Figure~\ref{fig:manifold:density}, we compare our model with the ground truth and find that we can successfully capture the manifold density.

\begin{table}[h]
\begin{center}
\begin{tabular}{ |p{3cm}|p{1cm}|p{2cm}|p{4.5cm}|  }
 \hline
Type of block &Number&Input shape&Affine coupling layer widths\\
 \hline
Fully connected & $2$   &$2$&   $1\rightarrow 24\rightarrow 24\rightarrow  24\rightarrow 1$\\
 \hline
\end{tabular}
\end{center}
\caption{The network structure for toy datasets. \label{tb:structure:toy}}

\begin{center}
\begin{tabular}{ |p{3cm}|p{1cm}|p{2cm}|p{4.5cm}|  }
 \hline
Type of block &Number&Input shape&Affine coupling layer widths\\
 \hline§
Fully connected & $6$   &$3$&  \makecell{ even: $2\rightarrow 24\rightarrow 24\rightarrow  24\rightarrow 1$\\
  \text{ }odd: $1\rightarrow 24\rightarrow 24\rightarrow  24\rightarrow 2$}\\
 \hline
\end{tabular}
\end{center}
\caption{The network structure for S-Curve dataset. If we denote the input vector of each coupling is $\rvx$, the function $s$ and $t$ takes $\rvx[0]$ in the first coupling layer and  $\rvx[1]$ in the second coupling layer. \label{tb:structure:scurve}}

\begin{center}
\begin{tabular}{ |p{3cm}|p{1cm}|p{2cm}|p{4.5cm}|  }
\hline
Type of block &Number&Input shape&Affine coupling layer widths\\
\hline
Downsampling    & 1 & $(1,32,32)$ & \\
Convolution     & 2 & $(4,16,16)$ & $2\rightarrow 16\rightarrow 16\rightarrow 4$ \\
Downsampling    & 1 & $(4,16,16)$ & \\
Convolution     & 2 & $(16,8,8)$  & $8\rightarrow16\rightarrow16\rightarrow 16$ \\
Flattening      & 1 & $(16,8,8)$  & \\
Fully connected & 2 & $1024$      & $512\rightarrow 24\rightarrow 24\rightarrow  24\rightarrow 512$ \\
\hline
\end{tabular}
\end{center}
\caption{The network structure for our fading square dataset experiments. If we denote the input vector of each coupling to be $\rvx$, the functions $s$ and $t$ take values $\rvx[1{:}2]$ in the even coupling layer and  $\rvx[3]$ in the odd coupling layer. 
\label{tb:structure:sauqre}}

\begin{center}
\begin{tabular}{ |p{3cm}|p{1cm}|p{2cm}|p{4.5cm}|  }
 \hline
Type of block &Number&Input shape&Affine coupling layer widths\\
 \hline
Downsampling    & 1 & $(1,28,28)$ & \\
Convolution     & 4 & $(4,14,14)$ & $2\rightarrow 16\rightarrow 16\rightarrow 4$\\
Downsampling    & 1 & $(4,14,14)$ & \\
Convolution     & 4 & $(16,7,7)$  & $8\rightarrow16\rightarrow16\rightarrow 16$\\
Flattening      & 1 & $(16,7,7)$  & \\
Fully connected & 2 & $784$       & $392\rightarrow 24\rightarrow 24\rightarrow  24\rightarrow 392$\\
 \hline
\end{tabular}
\end{center}
\caption{Network structure for synthetic MNIST dataset experiment. \label{tb:structure:mnist}}
\end{table}

\subsection{S-Curve dataset\label{app:scurve}}
To fit our model to the data, we use the Adam optimizer with learning rate $5\times 10^{-4}$, batch size $500$ and train the model for $200k$ iterations. We anneal the learning rate with a factor of $0.9$ every $10k$ iteration.

\subsection{Fading Square dataset \label{app:fading}}
To fit the data, we train our model for $20k$ iterations with a batch size of $100$ using the Adam optimizer. The learning rate is initialized to $5{\times}10^{-4}$ and decays with a factor of $0.9$ at every $1k$ iteration. We additionally use an $L2$ weight decay with factor $0.1$.  

\subsection{MNIST Dataset\label{app:mnist}}
\subsubsection{Implicit data generation model\label{app:mnist:implicit}}
To fit an implicit model to the MNIST dataset, we first train a Variational Auto-Encoder (VAE)~\cite{kingma2013auto} with Gaussian prior $p(\rvz){=}\mathcal{N}(0,I)$. The encoder is  $q(\rvz|\rvx)=\mathcal{N}(\mu_\theta(\rvx),\Sigma_\theta(\rvx))$ where $\Sigma$ is a diagonal matrix. Both $\mu_\theta$ and $\Sigma_\theta$ are parameterized by a 3-layer feed-forward neural network with a ReLU activation and the size of the two hidden outputs are $400$ and $200$. We use a Gaussian decoder $p(\rvx|\rvz)=\mathcal{N}(g_\theta(\rvz),\sigma_{\rvx}^2I)$ with fixed variance $\sigma_{\rvx}=0.3$. The $g_\theta$ is parameterized by a 3-layer feed-forward neural network with hidden layer sizes $200$ and $400$. The activation of the hidden output uses a ReLU and we utilize a Sigmoid function in the final layer output to constrain the output between 0 and 1. The training objective is to maximize the lower bound of the log-likelihood
\begin{equation*}
    \log p(\rvx)\geq \int q(\rvz|\rvx)\log p(\rvx|\rvz)dz- \KL(q(\rvz|\rvx)||p(\rvz)),
\end{equation*}
see \cite{kingma2013auto} for further details. We use an Adam optimizer with a learning rate $1\times 10^{-4}$ and batch size 100 to train the model for 100 epochs. After training, we sample from the model by first taking a sample $\rvz\sim p(\rvz)$ and letting $\rvx=g_\theta(\rvz)$. This is equivalent to taking a sample from an implicit model $p_\theta(\rvx)=\int \delta (\rvx-g_\theta(\rvz))d(\rvz)d\rvz$, see \cite{tolstikhin2017wasserstein} for further discussion regarding this implicit model construction. In Figure~\ref{fig:mnist:training:samples}, we plot samples from the trained implicit model with $\texttt{dim}(\rvz)=5$, $\texttt{dim}(\rvz)=10$ and the original MNIST data.

\subsubsection{Flow model training}
We train our flow models to fit the synthetic MNIST dataset with intrinsic dimensions 5 and 10 and the original MNIST dataset. In all models, we use the Adam optimizer with learning rate $5\times 10^{-4}$ and batch size $100$. We train the model for $300k$ iterations and, following the initial $100k$ iterations, the learning rate decays every $10k$ iterations by a factor $0.9$.

\begin{figure}[h]
\centering
\begin{subfigure}{0.3\textwidth}
\centering
\includegraphics[width=0.9\linewidth]{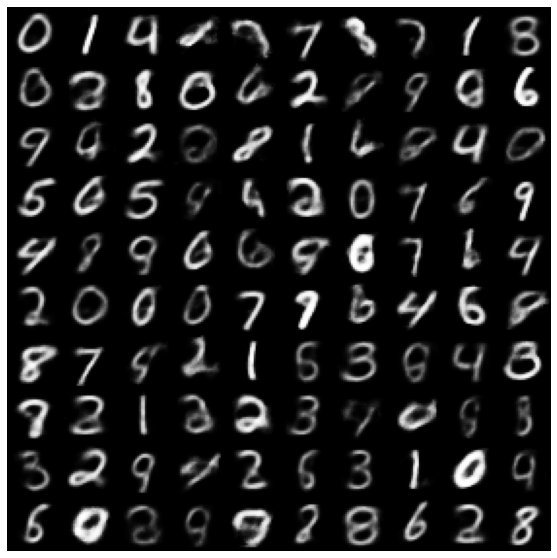} 
\caption{$\texttt{dim}(\rvz)=5$}
\end{subfigure}
\begin{subfigure}{0.3\textwidth}
\centering
\includegraphics[width=0.9\linewidth]{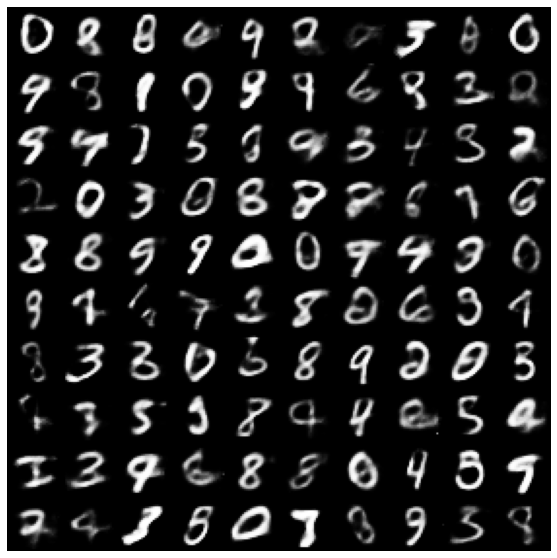} 
\caption{$\texttt{dim}(\rvz)=10$}
\end{subfigure}
\begin{subfigure}{0.3\textwidth}
\centering
\includegraphics[width=0.9\linewidth]{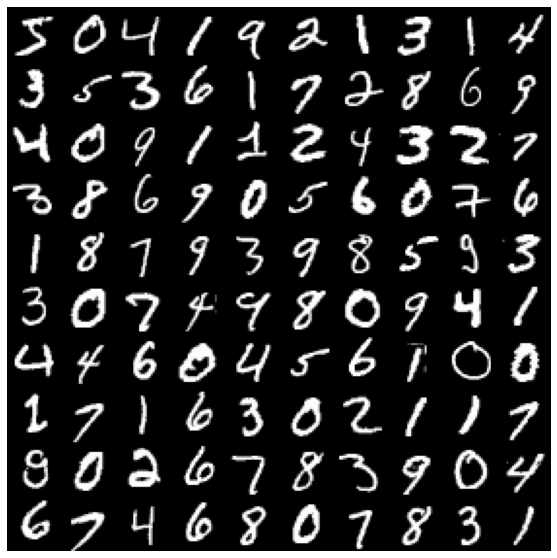}
\caption{MNIST Dataset}
\end{subfigure}
\caption{Training data for the flow model. Figure (a) and (b) are synthetic MNIST samples from two implicit models with latent dimension 5 and 10 respectively. Figure (c) are samples from the original MNIST dataset.\label{fig:mnist:training:samples}}

\centering
\begin{subfigure}{0.3\textwidth}
\centering
\includegraphics[width=0.9\linewidth]{img/mnist_sample_5.png} 
\caption{$\texttt{dim}(\rvz)=5$}
\end{subfigure}
\begin{subfigure}{0.3\textwidth}
\centering
\includegraphics[width=0.9\linewidth]{img/mnist_sample_10.png} 
\caption{$\texttt{dim}(\rvz)=10$}
\end{subfigure}
\begin{subfigure}{0.3\textwidth}
\centering
\includegraphics[width=0.9\linewidth]{img/mnist_samples.png}
\caption{MNIST}
\end{subfigure}
\caption{Samples from our methods. Figure (a) and (b) are samples flow models trained on synthetic MNIST data with intrinsic dimension 5 and 10 respectively. Figure (c) are samples from a flow model that trained on original MNIST data.\label{fig:mnist:flow:samples}}

\centering
\begin{subfigure}{0.3\textwidth}
\centering
\includegraphics[width=0.9\linewidth]{img/flow_mnist_5.png} 
\caption{$\texttt{dim}(\rvz)=5$}
\end{subfigure}
\begin{subfigure}{0.3\textwidth}
\centering
\includegraphics[width=0.9\linewidth]{img/flow_mnist_10.png} 
\caption{$\texttt{dim}(\rvz)=10$}
\end{subfigure}
\begin{subfigure}{0.3\textwidth}
\centering
\includegraphics[width=0.9\linewidth]{img/flow_mnist_o.png}
\caption{MNIST}
\end{subfigure}
\caption{Samples from traditional non-volume preserving flow models with fixed Gaussian prior. \label{fig:mnist:traditional:flow:samples}}
\end{figure}

In Figure \ref{fig:mnist:flow:samples}, we plot the samples from our models trained on three different training datasets. In Figure \ref{fig:mnist:traditional:flow:samples}, we also plot the samples from traditional flow models that were trained on these three datasets using the same experiment settings. We find that the samples from our models are 
comparably sharper than those from traditional flow models.

\subsection{Color image experiments\label{app:network:color}}
Our network for color image datasets is based on an open source implementation\footnote{\url{https://github.com/chrischute/glow}}, where we modify the Glow~\cite{kingma2018glow} structure to make it volume preserving. The volume preserving affine coupling layer is similar to the incompressible flow, however here the $s(\cdot)$ function takes the form:

\begin{align}
 s(\cdot)=\text{scale}*\text{tanh}(\mathrm{NN}(\cdot))
\end{align}

where `scale' are channel-wise learnable parameters and $\mathrm{NN}(\cdot)$ is the neural network. For the actnorm function \mbox{$\rvy_{ij}=\rvs\odot \rvx_{ij}+\rvb$}, where the log-determinant for position $\rvx_{ij}$ is $\textrm{sum}(\log|\rvs|)$, we normalize the $\widetilde{\log|\rvs|}=\log|\rvs|-\textrm{mean}(\log|\rvs|)$, such that $\textrm{sum}(\widetilde{\log|\rvs|})=0$. We use LU decomposition for the $1{\times}1$ convolution, where $W=PL(U + \text{diag}(\rvs))$ with log-determinant $\log|\det(W)|=\textrm{sum}(\log |\rvs|)$. We can then normalize $\widetilde{\log |\rvs|}=\log |\rvs|-\textrm{mean}(\log |\rvs|)$ to ensure it is volume preserving. All experiments utilise a Glow architecture consisting of four blocks of twenty affine coupling layers, applying the multi-scale architecture between each block. The neural network in each affine coupling layer contains three convolution layers with kernel sizes 3,1,3 respectively and ReLU activation functions. We use 512 channels for all hidden convolutional layers, the Adam optimizer with learning rate $1e^{-4}$ and a batch-size of $100$, for all of our experiments.

\begin{figure}[h]
\centering
\begin{subfigure}{0.32\linewidth}
\centering
\includegraphics[width=0.9\linewidth]{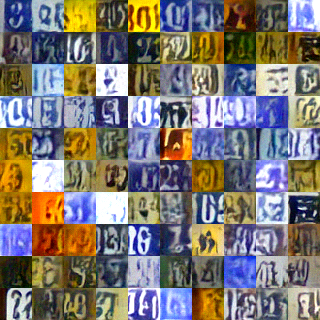}
\caption{SVHN Samples}
\end{subfigure}
\begin{subfigure}{0.32\linewidth}
\centering
\includegraphics[width=0.9\linewidth]{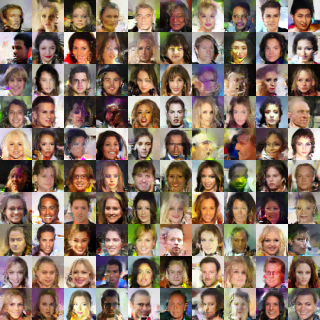}
\caption{CelebA Samples}
\end{subfigure}
\begin{subfigure}{0.32\linewidth}
\centering
\includegraphics[width=0.9\linewidth]{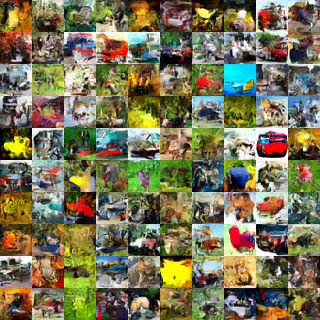}
\caption{CIFAR Samples}
\end{subfigure}

\begin{subfigure}{0.32\linewidth}
\centering
\includegraphics[width=0.9\linewidth]{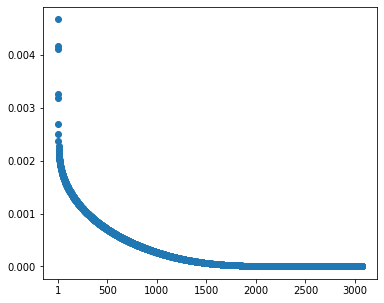}
\caption{SVHN Eigenvalues}
\end{subfigure}
\begin{subfigure}{0.32\linewidth}
\centering
\includegraphics[width=0.9\linewidth]{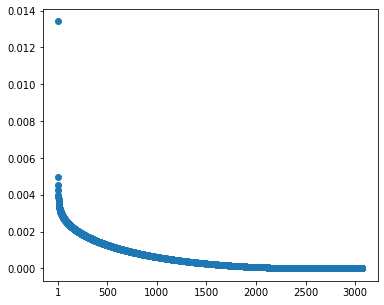}
\caption{CelebA Eigenvalues}
\end{subfigure}
\begin{subfigure}{0.32\linewidth}
\centering
\includegraphics[width=0.9\linewidth]{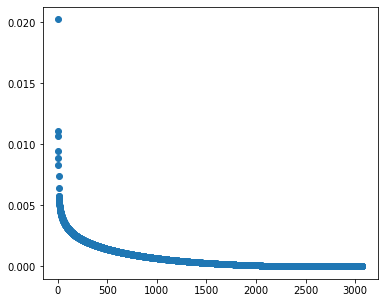}
\caption{CIFAR Eigenvalues}
\end{subfigure}

\caption{Samples and eigenvalues for three models. In each case we can find approximately half of the eigenvalues converge to $0$ but 
a large number of eigenvalues remain greater than $0$. \label{fig:cifar:color}}
\end{figure}

We conduct further experiments on real-world color images: SVHN, CIFAR10 and CelebA. Data samples from SVHN and CIFAR10 are represented by $32{\times}32{\times}3$ dimension vectors and for CelebA our pre-processing involves centre-cropping each image and resizing to $32{\times}32{\times}3$. Pixels take discrete values from $[0,\ldots,255]$ and we follow a standard dequantization process~\cite{uria2013rnade}: $x\nolinebreak=\nolinebreak\frac{x+\text{Uniform}(0,1)}{256}$ to transform each pixel to a continuous value in $[0,1]$. The manifold structure of these image datasets has been less well studied and may be considered more complex than the datasets previously explored in this work; each potentially lying on a union of disconnected and discontinuous manifolds. Similar to the MNIST experiment, we add small Gaussian noise (with standard deviation $\sigma_x{=}0.01$) for twenty initial training epochs in order to smooth the data manifold and then anneal the noise with a factor of $0.9$ for a further twenty epochs. Figure~\ref{fig:cifar:color} provides samples from the models that are trained on these three datasets. We find that our models can generate realistic samples 
and thus provide evidence towards the claim that our volume-preserving flow, with learnable prior model, does not limit the expressive power of the network. Experimentally, in comparison with the Glow model, we do not observe consistent sample quality improvement and conjecture that the intrinsic dimension mismatch problem, pertaining to these color image datasets, is less severe than the previously considered grey-scale images and constructed toy data. The data manifold has 
large intrinsic dimensionality due to high redundancy and stochasticity in the RGB channels~\cite{yu2010improved}. We then conjecture that color image data distributions typically have a relatively large number of intrinsic dimensions. This conjecture is consistent with recent generative model training trends: a latent variable model \eg VAE usually requires very large latent dimension to generate high-quality images \cite{vahdat2020nvae,child2020very}. For example, in the state-of-the-art image generation work\footnote{The latent dimension in GAN based models \cite{goodfellow2014generative,wgan1,gulrajani2017improved} are chosen to be small (\eg latent variable has dimension 128 for CIFAR10), but the mode-collapse phenomenon, commonly observed in GANs, indicates that those with small latent dimension will underestimate the support of the data distribution.} \cite{vahdat2020nvae}, a VAE with latent size $153600$ is used to fit a CIFAR10 dataset with shape $32{\times}32{\times}3$. These considerations lead us to believe that the estimation of intrinsic dimension for natural images forms an important future research direction. Towards tackling this, one proposal would involve using the results obtained from our model to inform the latent dimension of the latent variable models. We leave deeper study of the  geometry of image manifolds to future work.
\begin{table*}[t]
    \centering
        \caption{\label{fig:comp:cef} FID Comparisons of Samples. We report the FID values (lower indicates better quality) of  our model with $K=[512,2000,2500,3000]$. We surprisingly find that when we increase the dimension of the manifold, the FID becomes higher. We hypothesize that the manifold spanned by the leading eigenvectors can capture the images lying in the modes, resulting in a better FID value. We also compare our method to CEF \cite{ross2021tractable} model with manifold dimension $K=512$ and  find our method can consistently achieve better FID value.\label{tab:fid}}
    \begin{tabular}{c|c|c|c|c|c}
        \toprule
       Model  & CEF ($512$) & Ours ($512$) & Ours ($2000$) & Ours ($2500$) & Ours ($3000$) \\
       \midrule
        FID & 105.1 & \textbf{44.8} &54.4 & 62.8 &63.1 \\
        \bottomrule
    \end{tabular}
    \label{tab:my_label}
\end{table*}

\section{Empirical Verifications and Comparisons\label{app:comparison}}

\begin{figure}[h]
\centering
\begin{subfigure}{.24\textwidth}
\centering
\includegraphics[width=\textwidth]{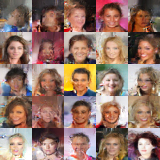}
\caption{Samples with $K=1000$}
\end{subfigure}
\begin{subfigure}{.24\textwidth}
\centering
\includegraphics[width=\textwidth]{new_img/2000_samples.png}
\caption{Samples with $K=2000$}
\end{subfigure}
\begin{subfigure}{.24\textwidth}
\centering
\includegraphics[width=\textwidth]{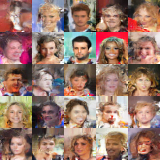}
\caption{Samples with $K=2500$}
\end{subfigure}
\begin{subfigure}{.24\textwidth}
\centering
\includegraphics[width=\textwidth]{new_img/3072_samples.png}
\caption{Samples with $K=3000$}
\end{subfigure}
\caption{Samples from the model $K$-dimensional manifolds prior.}
\label{fig:celebe:sample}
\end{figure}

We compare our model with the recently proposed CEF model~\cite{ross2021tractable}, which is also proposed to learn a flow on manifold distribution. Different from our model where different $K$ values can be selected after training  the model, in the CEF model, the intrinsic dimension of the model has to be pre-specified. We then choose $K=512$ and train the model on the CelebA ($32\times32$) dataset. We reuse their open-source code~\url{https://github.com/layer6ai-labs/CEF} and the $32\times32$ model architecture (see paper~\cite{ross2021tractable} for details). In Figure\ref{fig:comp:cef}, we compare the reconstructions and the samples from both models with the same manifold dimension $K=512$. We can find the CEF model has a better reconstruction quality than our models, we hypothesize that this is because our model is not trained with an intrinsic dimension $K=512$, so the reconstruction quality cannot be guaranteed when we use the latent code with dimension 512. However, we observe our model achieves better sample quality (see Table\ref{tab:fid} for a FID comparison) compared to CEF.
This phenomenon (good reconstruction but bad sample quality) is similar to the latent space mismatch problem  in the latent variable models \cite{dai2019diagnosing,zhao2017infovae}, where the aggregate distribution $q_\phi(z)=\int \delta (z-g(x))\mathbb{P}_d(x)$ mismatches the prior $p(z)$, ($g$ is the inverse flow function). Since our model can learn the prior and the manifold prior with $K=512$ is constructed by the leading eigen-vectors, it can alleviate the prior mismatch problem and has better sample quality.

\begin{figure}[h]
\centering
\begin{subfigure}{\textwidth}
\centering
\includegraphics[width=\textwidth]{new_img/true.png}
\caption{Reconstruction Ground Truth}
\end{subfigure}
\begin{subfigure}{\textwidth}
\centering
\includegraphics[width=\textwidth]{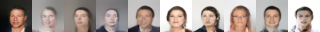}
\caption{Reconstruction from our model with $K=512$}
\end{subfigure}
\begin{subfigure}{\textwidth}
\centering
\includegraphics[width=\textwidth]{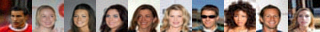}
\caption{Reconstruction from CEF with $K=512$}
\begin{subfigure}{.49\textwidth}
\centering
\includegraphics[width=\textwidth]{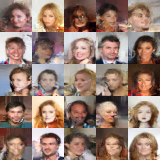}
\caption{Samples from our model with $K=512$}
\end{subfigure}
\begin{subfigure}{.49\textwidth}
\centering
\includegraphics[width=\textwidth]{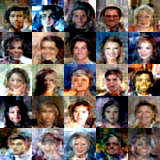}
\caption{Samples from CEF with $K=512$}
\end{subfigure}
\end{subfigure}

\caption{Comparisons with CEF model. We compare the reconstructions (a,b,c) and samples (d,c) from our model and CEF model with the same manifold dimension ($K=512$), we can see CEF has a better reconstruction quality whereas our model has a better sample quality, also see \ref{tab:fid} for  FID comparisons and main text for a discussion of the phenomenon.}
\end{figure}

\section{License\label{app:license}}

The CelebA \cite{liu2015faceattributes} and SVHN \cite{netzer2011reading} dataset is available for non-commercial research purposes only. We didn't find licenses for CIFAR \cite{krizhevsky2009learning} and MNIST \cite{lecun1998mnist}. The CelebA dataset may contain personal identifications.

The code \url{https://github.com/chrischute/glow} we made use of is under the MIT License.

\end{document}